\documentclass[lettersize,journal]{IEEEtran}
\usepackage{amsmath,amsfonts}
\usepackage{algorithmic}
\usepackage{algorithm}
\usepackage{array}
\usepackage[caption=false,font=normalsize,labelfont=sf,textfont=sf]{subfig}
\usepackage{textcomp}
\usepackage{stfloats}
\usepackage{url}
\usepackage{verbatim}
\usepackage{graphicx}
\usepackage{cite}
\usepackage{Jiazhi_style}
\usepackage{orcidlink}

\begin{document}

\title{A Critical Review of Predominant Bias \\in Neural Networks}

\author{Anonymous Submission}

\author{Jiazhi Li\orcidlink{0000-0003-3938-7989}, Mahyar Khayatkhoei\orcidlink{0000-0002-7326-861X}, Jiageng Zhu\orcidlink{0009-0002-0162-6534}, Hanchen Xie\orcidlink{0009-0004-4474-4877}, \\ Mohamed E. Hussein\orcidlink{0000-0002-4707-9313}, \textit{Member, IEEE}, and Wael AbdAlmageed\orcidlink{0000-0002-8320-8530}, \textit{Member, IEEE}

\thanks{Wael AbdAlmageed is affiliated with Holcombe Department of Electrical And Computer Engineering at Clemson University (e-mail: wabdalm@clemson.edu). All other authors are affiliated with USC Information Sciences Institute (e-mail: jiazhil@usc.edu; khayatkh@usc.edu; jiagengz@usc.edu; hanchenx@usc.edu; mehussein@isi.edu).
Jiazhi Li and Jiageng Zhu are also at USC Ming Hsieh Department of Electrical and Computer Engineering.
Hanchen Xie is also at USC Thomas Lord Department of Computer Science.
Mohamed E. Hussein is also at Alexandria University.}}

\IEEEpubid{}

\maketitle

\begin{abstract}

Bias issues of neural networks garner significant attention along with its promising advancement.
Among various bias issues, mitigating two predominant biases is crucial in advancing fair and trustworthy AI: (1) ensuring neural networks yields even performance across demographic groups, and (2) ensuring algorithmic decision-making does not rely on protected attributes.
However, upon the investigation of \pc papers in the relevant literature, we find that there exists a persistent, extensive but under-explored confusion regarding these two types of biases.
Furthermore, the confusion has already significantly hampered the clarity of the community and subsequent development of debiasing methodologies.
Thus, in this work, we aim to restore clarity by providing two mathematical definitions for these two predominant biases and leveraging these definitions to unify a comprehensive list of papers. 
Next, we highlight the common phenomena and the possible reasons for the existing confusion.
To alleviate the confusion, we provide extensive experiments on synthetic, census, and image datasets, to validate the distinct nature of these biases, distinguish their different real-world manifestations, and evaluate the effectiveness of a comprehensive list of bias assessment metrics in assessing the mitigation of these biases.
Further, we compare these two types of biases from multiple dimensions including the underlying causes, debiasing methods, evaluation protocol, prevalent datasets, and future directions.
Last, we provide several suggestions aiming to guide researchers engaged in bias-related work to avoid confusion and further enhance clarity in the community.

\end{abstract}


\section{Introduction}
\label{sec:introduction}

\begin{figure*}[h!]
\centering
  \includegraphics[width=1\linewidth]{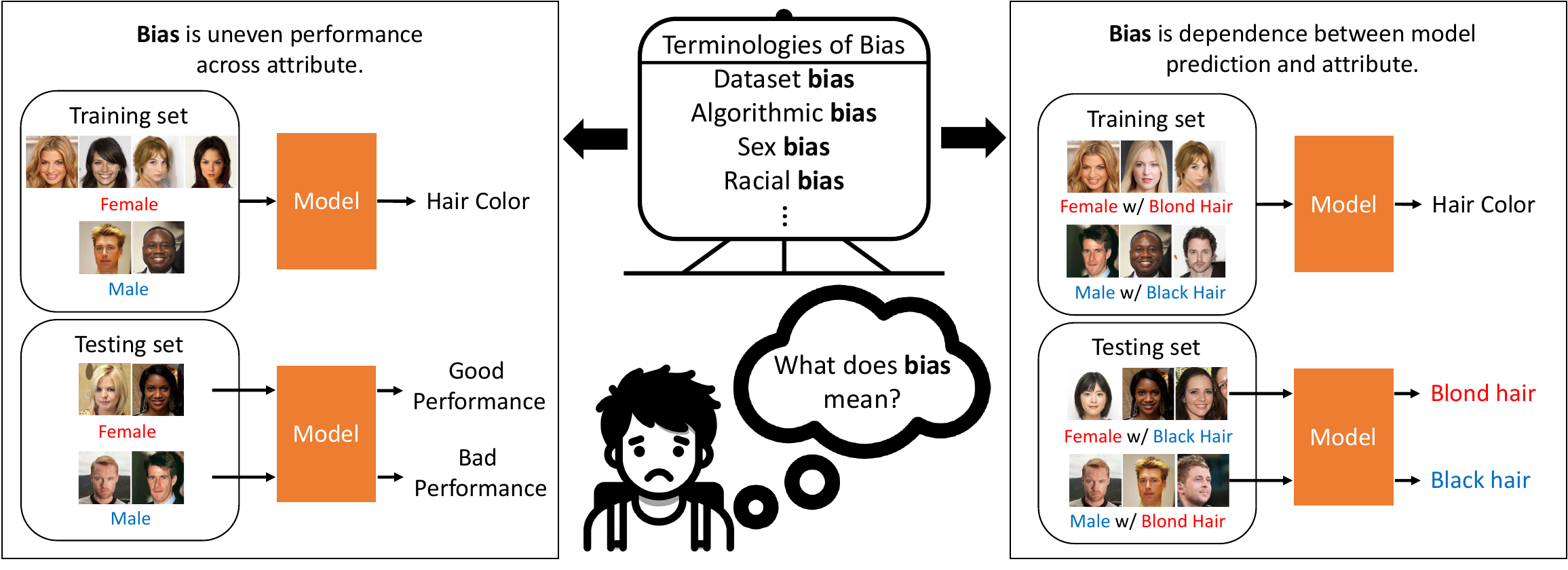}
  \caption{The same set of terminology about bias is interpreted differently by experts, which significantly confuses the understanding of the audience. By investigating \pc papers about prevalent bias issues, we discover that there exists significant confusion regarding these prevalent bias issues. The confusion is evident in several ways such as ambiguity of terminology, inaccurate motivation, and lack of terminology reuse.
  Most notably, several studies inaccurately motivate themselves on a particular bias while actually addressing a different type of bias. 
  This prevailing confusion considerably impedes the clarity of related work. 
  Thus, we propose new definitions to unify the existing literature and pave a clear path for future research.}
\label{fig:idea}
\end{figure*}

\begin{table*}[b]
\caption{Main distinctions between Type I Bias and Type II Bias.}
\label{tab:teaser}
\centering

\begin{tabular}{lcc}
\toprule
                        & Type I Bias                                          & Type II Bias                                                       \\
                        \midrule
Manifestation           & Uneven performance across attribute $A$              & Dependence between model prediction $\hat{Y}$ and attribute $A$    \\
Use of ground truth $Y$              & \cmark                                               & \xmark                                                             \\
\multirow{2}{*}{Representative example} & Facial recognition systems exhibit lower performance & Bank loan systems tend to approve loans more frequently            \\
                        & in one demographic group compared to others         & for one demographic group compared to others                      \\
Possible reason         & Insufficient training in underrepresented group     & Correlation between the target $Y$ and the attribute $A$ in training set \\
\bottomrule
\end{tabular}

\end{table*}

\IEEEPARstart{N}{eural} networks have shown promising advances in many prediction and classification tasks~\cite{imagenet,ResNet,atari_DRL}.
Along with the impressive capability of neural networks, its societal impact has garnered great attention~\cite{Timnit_sex_classification_PPB, CSAD, MvCoM}, particularly regarding \emph{protected attributes} (\eg sex, race, and age), which cannot be used in the decision-making process~\cite{protected_attributes}.
Failing to carefully consider protected attributes while deploying neural networks can lead to bias issues and severely compromise fairness for specific demographic groups in various real-world applications~\cite{Timnit_sex_classification_PPB, COMPAS, diabetes_chapter}.
For instance, facial recognition systems may more correctly recognize males than females~\cite{DebFace}.
\rev{Besides, Artificial Intelligence-assisted bank loan systems may classify a higher proportion of male applicants as having bad credit than female applicants~\cite{CSAD}.}

\rev{The underlying bias issues of neural networks, involved in the aforementioned examples, lead to important discussions~\cite{CSAD, MvCoM, DebFace, RFW_IMAN, RL_RBN, GAC,BlindEye_IMDB_eb,learn_not_to_learn_Colored_MNIST,DI, LfF_CelebA_Bias_conflicting,End,BCL}.
Specifically, these aforementioned examples highlight the presence of two distinct prevalent types of biases.
Without loss of generality, for disambiguation, these two predominate biases can be summarized as follows:

\begin{itemize}
    \item The model yields uneven performance across different demographic attributes, referred to as \emph{Type I Bias}.
    \item The model depends on demographic attributes to make predictions, referred to as \emph{Type II Bias}. 
\end{itemize}}
\noindent

\rev{Although these two prevalent types of biases differ in many aspects, as highlighted in~\cref{tab:teaser}, the current literature often ambiguously groups them under the general term ``bias" (\eg dataset bias, algorithmic bias, sex bias, or racial bias)~\cite{BlindEye_IMDB_eb, Back_MI, FairCal} and interpret them differently across scenarios.
Furthermore, numerous works addressing one type of bias inadvertently cite the other as their motivation~\cite{RFW_IMAN,RL_RBN,FairCal}.
Additionally, the taxonomy of bias issues in existing survey papers may not sufficiently distinguish between them or explicitly acknowledge their differences~\cite{MLbias_survey,causal_based_fairness,discussion_on_DP_EO}.}

\rev{Overlooking the distinction between these two types of biases significantly compromises clarity in the current literature and leads to various negative consequences.
Specifically, for new researchers, the lingering question of which specific type of bias a paper addresses creates unnecessary confusion. 
Furthermore, the widespread confusion surrounding these biases and the lack of clear definitions to separate them results in weak motivation, ambiguous statements, and vague contributions in the existing debiasing work, significantly impeding the clarity of the associated research.
Additionally, persistent conflation of these biases, usage of inappropriate references, and unfair comparison between methods addressing different biases can lead to an expanding misunderstanding over time.
Besides, this confusion complicates the resolution of bias issues and hinders the advancement of future work in this field.}

To that end, the main goal of this paper is to unify the existing literature about Type I Bias and Type II Bias, rectify the common confusion regarding them, and alleviate the cognitive burden for future research.
The contributions of this paper can be summarized as follows:
\begin{itemize}
    \item Proposing General mathematical definitions for Type I Bias and Type II Bias (\cref{sec:definition}) and providing a summary of their corresponding related work (\cref{sec:following}). These can be utilized as a roadmap for future work.

    \item Unifying a comprehensive list of work and relevant fairness criteria under the definition of Type I Bias and Type II Bias (\cref{sec:unifying}).
    
    \item Elucidating the existing phenomena stemming from the confusion between Type I Bias and Type II Bias (\cref{sec:confusion}), and exploring the underlying reasons that contribute to the confusion (\cref{sec:reason}).

    \item Conducting extensive experiments to examine the distinction between Type I Bias and Type II Bias (\cref{sec:distinction}).

    \item Offering some suggestions to foster a clear community regarding these bias issues (\cref{sec:suggestion}).
    
\end{itemize}

\section{Definitions}
\label{sec:definition}

To define and distinguish these two types of biases, we first establish several key concepts. Given a dataset $\mathcal{D}: {\mathcal{X}, \mathcal{Y}, \mathcal{A}}$ consisting of instances ${x, y, a}$ where each sample $x \in \mathcal{X}$ is annotated with an attribute label $a$ (e.g., sex) and a ground truth label $y$ for a specific downstream task (e.g., identity in face recognition), the model $f: \mathcal{X} \to \mathcal{Y}$ takes $x$ as input and outputs the predicted label $\hat{y}$.
In this section, we introduce formal mathematical definitions for these two types of biases, referred to as Type I Bias and Type II Bias, which will be consistently used throughout the paper. In the following sections, we will review \pc papers to demonstrate that various commonly discussed bias issues can be unified using these definitions and explore the phenomena and reasons behind the existing confusion between these bias issues.

\subsection{Type I Bias}

\rev{The manifestation of Type I Bias is uneven model performance across different demographic groups~\cite{RFW_IMAN, RL_RBN, DebFace, GAC, MvCoM}.}
Specifically, model performance can be evaluated using various metrics, \eg error rate~\cite{Timnit_sex_classification_PPB,fairnessgan_DP_difference_error_rate}, loss~\cite{representation_disparity}, accuracy~\cite{multiaccuracy}, average precision (AP)~\cite{DP_difference_fpr_GAN_debiasing}, positive predictive value (PPV), 
true positive rate (TPR)~\cite{pass,BR_Net_dataset_vs_task}, false positive rate (FPR)~\cite{FPR_Penalty_Loss}, average false rate (AFR), mean AFR (M AFR)~\cite{inclusivefacenet}, confusion matrix~\cite{DebFace}, F1 score~\cite{BR_Net_dataset_vs_task}, receiver operating characteristic curve (ROC)~\cite{RL_RBN, fairnessgan_DP_difference_error_rate, SAN, Asymmetric_Rejection_Loss,debias_balanced_AUCROC}, area under the ROC (AUC)~\cite{FlowSAN, DebFace, BR_Net_dataset_vs_task}.
All these metrics can be unified under the format of a distance measure $d(\hat{Y},Y)$, evaluated based on model prediction $\hat{Y}$ and ground truth label $Y$.
Thus, we can formally define this type of bias as follows:
\begin{definition}
    \label{def:Type_I_Bias}
Type I Bias. A model $f$ involves \emph{Type I Bias} if $f$ yields uneven performance $d(\hat{Y},Y)$ across attribute $A$,
\begin{align}
    \sup_{a,a'\in \mathcal{A}, d \in \mathcal{M}} \abs{d(\hat{Y},Y|A=a) - d(\hat{Y},Y|A=a')} > 0
\end{align}
\end{definition}
\noindent
where $a,a'$ are possible values of $A$ (\eg female and male), and $\mathcal{M}$ is the set of all potential performance metrics.

\subsection{Type II Bias}

\rev{On the other hand, the manifestation of Type II Bias is dependence between model prediction and attribute~\cite{BlindEye_IMDB_eb,learn_not_to_learn_Colored_MNIST,DI, LfF_CelebA_Bias_conflicting,End,CSAD,BCL}.}
Specifically, these attributes can be categorized by sensitive/protected attributes~\cite{Fairalm_DP_difference_false_positive_rate, GDP} (\eg sex in creditworthiness prediction) or spurious attributes~\cite{Group_DRO,JTT} (\eg texture in object recognition).
Both of these scenarios can be unified as the dependence between model prediction and the specific attribute.
Thus, we can formally define this type of bias as follows:
\begin{definition}
    \label{def:Type_II_Bias}
Type II Bias. A model $f$ involves \emph{Type II Bias} if model prediction $\hat{Y}$ is not independent with attribute $A$,
\begin{align}
\sup_{a,a' \in \mathcal{A}} \abs{P(\hat{Y}|A=a) - P(\hat{Y}|A=a')} > 0 
\end{align}
\end{definition}

\noindent
where $a,a'$ are possible values of $A$ (\eg female and male).

\section{Method}
\label{sec:method}

In this section, we introduce the method used to conduct the investigation on a set of \pc papers that discuss relevant bias issues.
Specifically, to construct the initial set of relevant work, we search the keywords ``bias" or ``fair" in the title of papers from NeurIPS, ICML, ICLR and FAccT published before February 2025. 
We include papers that discuss bias issues whose manifestation aligns with either Type I Bias or Type II Bias (we will detail the unification in~\cref{sec:unifying}).
We exclude papers that address other bias issues such as inductive bias~\cite{baxter2000model,zietlow2021demystifying}, implicit bias~\cite{fitzgerald2017implicit,camuto2021asymmetric}, selection bias~\cite{hernan2004structural,akbari2021recursive}, sampling bias~\cite{winship1992models,xu2022alleviating}, spectral bias~\cite{fang2024addressing}, exposure bias~\cite{li2024alleviating} or bias-variance~\cite{ha2024fine, chen2024on}.
Furthermore, to ensure we do not overlook any relevant papers without these keywords or from other prominent conferences such as CVPR, ICCV, and ECCV, we manually traversal the citation graph of the paper in the initial set and append the relevant papers that are either cited by or cite the papers in the initial set.

Once we identify the scope of the investigated papers, we read these papers to determine which type of bias they address by examining two aspects: problem statement and evaluation protocol.
We will elaborate on the criterion for categorizing papers into our definitions in~\cref{sec:unifying}.
To accommodate the recent emerging direction of addressing unlabeled and unknown bias, we enrich the taxonomy with an additional dimension about the status of attribute $A$.
As shown in~\cref{tab:taxonomy}, we count the number of papers in each category. 
Note that the total number is not equal to \pc since some papers address both types of biases.
We present the categorization list of all \pc investigated papers in Appendix.

\begin{table}[htbp]
\caption{The taxonomy of bias issues based on \pc papers.}
\label{tab:taxonomy}
\centering
\resizebox{0.45\textwidth}{!}{%

\begin{tabular}{lcccc}
\toprule
\multirow{2}{*}{Type of Bias} & \multicolumn{2}{c}{Attribute $A$} & \multirow{2}{*}{Papers} & \multirow{2}{*}{Examples}                                                   \\
\cmidrule(lr){2-3} 
                              & Known           & Labeled         &                         &                                                                             \\
                              \midrule
\multirow{3}{*}{Type I Bias}  & \cmark          & \cmark          & 253                     & \cite{DebFace,GAC,RL_RBN}                                                   \\
                              & \cmark          & \xmark          & -                       & -                                                                           \\
                              & \xmark          & \xmark          & -                       & -                                                                           \\
                              \midrule
\multirow{3}{*}{Type II Bias} & \cmark          & \cmark          & 246                     & \cite{learn_not_to_learn_Colored_MNIST,CSAD,End}                            \\
                              & \cmark          & \xmark          & 8                       & \cite{HEX_texture_bias1, ReBias_texture_bias2,rubi} \\
                              & \xmark          & \xmark          & 30                      & \cite{LfF_CelebA_Bias_conflicting,ECS,UBNet}                               \\
                              \midrule
Survey                        & -               & -               & 25                       & \cite{MLbias_survey,prediciton_quality_disparity,discussion_on_DP_EO}      \\
\bottomrule
\end{tabular}
}

\end{table}

\section{Unification}
\label{sec:unifying}

In this section, we clarify how bias issues discussed in existing literature align with our proposed definitions.
Generally, we categorize the bias into a specific type of bias in our definition if the presence of this bias implies the existence of bias in our definitions.
Furthermore, the categorization primarily relies on two key factors: the manifestation of bias issues explicitly addressed (if stated in ``Problem Statement" section) and the characteristics of evaluation protocol\footnote{For instance, Type I Bias involves training sets which yield the long-tail distribution, while Type II Bias typically involves training sets which yields the association between target label and attribute label.}.
Other aspects such as motivation, related work, method, or bias assessment are considered secondary factors for categorization. This is because certain papers, despite addressing different manifestations of bias, can exhibit similarities in these aspects, thereby leading to the confusion between these two types of biases, as elaborated in~\cref{sec:confusion}.

\subsection{Type I Bias}
The general form of Type I Bias is characterized by the uneven performance of the target across attributes. This definition can be extended to unify a wide range of papers by specifying the usage of performance metrics and the kind of target.
To clarify, several representative descriptions are shown as follows, \eg

\begin{itemize}
    \item \emph{``Racial bias indeed degrades the fairness of recognition system and the error rates on non-Caucasians are usually much higher than Caucasians."}~\cite{RL_RBN}
    \item \emph{``A certain demographic group can be better recognized than other groups."}~\cite{GAC}
    \item \emph{``Recognition accuracies depend on demographic cohort."}~\cite{RFW_IMAN}
\end{itemize}

\noindent
By specifying how performance is evaluated, Type I Bias covers a broad range of papers where model performance is evaluated using various criteria such as error rate~\cite{fairnessgan_DP_difference_error_rate}, loss~\cite{representation_disparity}, accuracy~\cite{multiaccuracy}, True Positive Rate (TPR)~\cite{pass}, False Positive Rate (FPR)~\cite{FPR_Penalty_Loss}, Receiver Operating Characteristic curve (ROC)~\cite{debias_balanced_AUCROC}, and Area Under the Curve (AUC)~\cite{DebFace}.
Furthermore, by specifying the kind of target, this definition can unify a wider range of papers.
For instance, considering sex as an attribute, the targets can include identity~\cite{DebFace, FairCal} (\eg face recognition), the attribute itself~\cite{Timnit_sex_classification_PPB,Fairface} (\eg sex classification), or other targets associated with protected attribute~\cite{minority_group_vs_sensitive_attribute,representation_disparity} (\eg facial attribute classification).
It is noteworthy that Type I Bias is predominantly discussed in various biometrics tasks~\cite{von_Mises_Fisher,FR_inherent_bias,SensitiveNets}. 
Compared with various types of targets, protected attributes (\eg sex, race, and age) are mainly considered the term of attribute in Type I Bias.

\subsection{Type II Bias}
\label{subsec:Type_II_Bias}
The general form of Type II Bias is characterized by the dependence between model prediction and attribute. 
This definition can be used to unify a broad spectrum of papers by considering the status of attribute and the kind of attribute.
The status of attribute is categorized into three groups, including known and labeled, known but unlabeled, and unknown.
Specifically, for known and labeled bias, several methods directly leverage attribute labels to explicitly apply supervision signal for bias mitigation~\cite{CSAD}.
For known but unlabeled bias, several methods mainly utilize the domain knowledge of specific bias attribute to design the module tailored for this bias attribute~\cite{HEX_texture_bias1}.
For unknown bias, several methods identify and emphasize bias-conflicting samples (those exhibiting the opposite bias present in the training set) to mitigate bias~\cite{ECS}. On the other hand, the kind of attribute mainly encompasses sensitive/protected attributes~\cite{machine_bias,annotation_bias,discrimination_score} and spurious attributes~\cite{LfF_CelebA_Bias_conflicting,Group_DRO,ECS}.
In the case of sensitive attributes, the reliance on them leads to a disproportionate assignment of specific predictions to particular demographic groups, thereby resulting in unfair treatment.
In this category, demographic parity~\cite{fairness_through_awareness}, a well-known fairness criterion, is often served as a debiasing objective. 
We present several representative descriptions as follows, \eg

\begin{itemize}
    \item \emph{``Demographic parity, which is satisfied when the predictions are independent of the sensitive attributes."}~\cite{DP_FFVAE}
    \item \emph{``Data fairness can be achieved if the generated decision has no correlation with the generated protected attribute."}~\cite{fairgan}
    \item \emph{``Ensuring that the positive outcome is given to the two groups at the same rate."}~\cite{LAFTR}
\end{itemize}

\noindent
In the case of spurious attributes, depending on them for decision-making will simplify the training process since models may utilize them as shortcut features instead of learning more comprehensive features during training. However, this leads to model predictions heavily relying on these attributes and further poor generalization performance in real-world applications since such spurious correlation between target and attribute does not generally exist. 
Several representative descriptions are shown as follows, \eg

\begin{itemize}
    \item \emph{``If bias features are highly correlated with the object class in the  dataset, models tend to use the bias as a cue for the prediction."}~\cite{BCL}
    \item \emph{``Since there are correlations between the target task label and the bias label, the target task is likely to rely on the bias information to fulfill its objective."}~\cite{CSAD}
    \item \emph{``If biased data is provided during training, the machine perceives the biased distribution as meaningful information."}~\cite{learn_not_to_learn_Colored_MNIST}
\end{itemize}

\begin{table*}[htbp]
\centering
   \caption{The summary of representative fairness criteria.}
   \label{tab:fairness_criteria}
   \resizebox{0.98\textwidth}{!}{%
   \begin{tabular}{lllc}
   \toprule
Category                                   & Notion             & Definition                                                                & Examples                                                 \\
\midrule
\multirow{3}{*}{Fairness \wrt Type I Bias} & Equalized odds~\cite{EO_define}       & $P(\hat{Y}=y_1|A=a_0,Y=y) = P(\hat{Y}=y_1|A=a_1,Y=y), y \in \{y_0,y_1\}$ &    \cite{FSCL,FURL_PS,von_Mises_Fisher}                                      \\ 
                                           & Equal opportunity~\cite{EO_define}    & $P(\hat{Y}=y_1|A=a_0,Y=y_1) = P(\hat{Y}=y_1|A=a_1,Y=y_1)$                &  \cite{CGL,POCAR,FATDM}                                          \\
                                           & Accuracy parity~\cite{Accuracy_parity}    & $P(\hat{Y}=Y|A=a_0) = P(\hat{Y}=Y|A=a_1)$                                &      \cite{multiaccuracy,disparate_mistreatment_on_FPR,Accuracy_parity}                                \\
                                           \midrule
Fairness \wrt Type II Bias                 & Demographic parity~\cite{fairness_through_awareness, counterfactual_fairness} & $P(\hat{Y}|A=a_0) = P(\hat{Y}|A=a_1)$                                    & \cite{DP_FFVAE,fairgan,decaf} \\
\bottomrule
\end{tabular}
}
\end{table*}

\subsection{Fairness Criteria}
Besides the papers that explore bias issues directly from the perspective of bias itself, there is another group of papers that leverage established fairness criteria (\eg demographic parity and equalized odds) as their debiasing objectives.
In this section, we first adopt the corresponding definitions of fairness from the definition of bias in~\cref{def:Type_I_Bias,def:Type_II_Bias}, and then demonstrate that relevant papers based on established fairness criteria can be categorized under these definitions.
Given that fairness is the opposite of bias, we can derive the fairness definition for each type of bias as follows,

\begin{definition}
    \label{def:Fairness_Type_I_Bias}
Fairness \wrt Type I Bias. A model $f$ is fair \wrt \emph{Type I Bias} if $f$ yields even performance $d(\hat{Y},Y)$ across attribute $A$, \ie
\begin{align}
    \sup_{a,a'\in \mathcal{A}, d \in \mathcal{M}} \abs{d(\hat{Y},Y|A=a) - d(\hat{Y},Y|A=a')} = 0
\end{align}
\end{definition}
\noindent
where $a,a'$ are possible values of $A$ (\eg female and male), and $\mathcal{M}$ is the set of all potential performance metrics.

\begin{definition}
    \label{def:Fairness_Type_II_Bias}
Fairness \wrt Type II Bias. A model $f$ is fair \wrt \emph{Type II Bias} if model prediction $\hat{Y}$ is independent with attribute $A$, \ie
\begin{align}
\sup_{a,a' \in \mathcal{A}} \abs{P(\hat{Y}|A=a) - P(\hat{Y}|A=a')} = 0 
\end{align}
\end{definition}
\noindent
where $a,a'$ are possible values of $A$ (\eg female and male).

Fairness criteria can be categorized into two key classes: group fairness and individual fairness~\cite{MLbias_survey, causal_based_fairness,discussion_on_DP_EO}.
Specifically, group fairness is founded on the idea that ``groups of people may face biases and unfair decisions", whereas individual fairness is grounded in the principle that ``similar individuals should receive similar decisions"~\cite{discussion_on_DP_EO}.
We mainly unify group fairness into our definitions since group fairness is more commonly used in fairness research~\cite{prediciton_quality_disparity}.
Group fairness encompasses several well-known fairness criteria such as demographic parity/statistical parity~\cite{fairness_through_awareness, counterfactual_fairness}, equalized odds/equality of odds~\cite{EO_define}, equal opportunity/equality of opportunity~\cite{EO_define}, and accuracy parity~\cite{Accuracy_parity}. The categorization of them under our fairness definitions is shown in~\cref{tab:fairness_criteria}.
Specifically, demographic parity, which requires $P(\hat{Y}|A=a_0) = P(\hat{Y}|A=a_1)$, is consistent with~\cref{def:Fairness_Type_II_Bias} when attribute $A$ is binary.
Equalized odds, which requires that both even true positive rate (TPR) ($P(\hat{Y}=y_1|Y=y_1)$) and even false positive rate (FPR) ($P(\hat{Y}=y_1|Y=y_0)$) across $A$, and equal opportunity, which is the weaker notion of equalized odds that focuses solely on the advantaged outcome where $Y=y_1$, align with~\cref{def:Fairness_Type_I_Bias} since TPR and FPR are included in the set of performance metrics $\mathcal{M}$.
Accuracy parity, where accuracy is represented by $P(\hat{Y}=Y)$, also aligns with~\cref{def:Fairness_Type_I_Bias} since accuracy is the element of $\mathcal{M}$.

\subsection{Summary}

Having unified the prevalent bias issues and well-known fairness criteria under our definitions, in this section, we summarize the main advantages of the proposed definitions.
First, the proposed definitions focus on the manifestation of predominant bias, which is more clear and easier to apply compared to definitions based on causes, since causes of these biases are debatable in some cases~\cite{BR_Net_dataset_vs_task, minority_group_vs_sensitive_attribute,RL_RBN}.
Second, the proposed definitions yield the general form, and by specifying the components in the general form, they can be used to unify a comprehensive list of papers, as summarized in~\cref{tab:unification}. 
Third, the proposed definitions, as the first definition to formally define dominant biases, bridge the gap between numerous fairness definitions~\cite{EO_define,counterfactual_fairness,fairness_through_awareness,fairness_under_unawareness,process_fairness_FPR_difference,impossibility_for_fair_repesentations,Accuracy_parity} and the significant shortage of formal bias definition.
Furthermore, compared with fairness definitions, bias definitions are more practical since encountering bias issues is more common in real-world scenarios, whereas achieving fairness, often considered an ideal benchmark, is rare in practice.
Fourth, given that the proposed bias definitions are relatively general, the corresponding fairness definitions are strict, hence aligning with the need for fairness as an ideal standard. 
Additionally, several well-known fairness criteria can be unified under the proposed fairness definitions.

\begin{table*}[htbp]
\centering
   \caption{The overview of the literature regarding Type I Bias and Type II Bias.}
   \label{tab:unification}

   \resizebox{0.9\textwidth}{!}{%

\begin{tabular}{ll|llc}
\toprule
Category                      & Description                                                        & \multicolumn{2}{l}{Subsettings}                                                                  & \multicolumn{1}{c}{Examples}                                                                               \\
\midrule
\multirow{14}{*}{Type I Bias} & \multirow{14}{*}{Uneven performance of target across attribute}    & \multirow{11}{*}{How is performance evaluated?}  & Error rate                                    & \cite{Timnit_sex_classification_PPB,fairnessgan_DP_difference_error_rate}                                  \\
                              &                                                                    &                                                  & Loss                                          & \cite{representation_disparity}                                                                            \\
                              &                                                                    &                                                  & Accuracy                                      & \cite{multiaccuracy}                                                                                       \\
                              &                                                                    &                                                  & Average precision                             & \cite{DP_difference_fpr_GAN_debiasing}                                                                     \\
                              &                                                                    &                                                  & True positive rate                            & \cite{pass,BR_Net_dataset_vs_task}                                                                         \\
                              &                                                                    &                                                  & False positive rate                           & \cite{FPR_Penalty_Loss}                                                                                    \\
                              &                                                                    &                                                  & Mean average false rate                       & \cite{inclusivefacenet}                                                                                    \\
                              &                                                                    &                                                  & Confusion matrix                              & \cite{DebFace}                                                                                             \\
                              &                                                                    &                                                  & F1 score                                      & \cite{BR_Net_dataset_vs_task}                                                                              \\
                              &                                                                    &                                                  & Receiver operating characteristic curve (ROC) & \cite{SAN, Asymmetric_Rejection_Loss,debias_balanced_AUCROC} \\
                              &                                                                    &                                                  & Area under the ROC (AUC)                & \cite{FlowSAN, DebFace, BR_Net_dataset_vs_task}                                                            \\
                             \cmidrule{3-5} 
                              &                                                                    & \multirow{3}{*}{Type of target}                  & Identity                                      & \cite{RL_RBN, RFW_IMAN, GAC}                                                                               \\
                              &                                                                    &                                                  & Attribute itself                              & \cite{Timnit_sex_classification_PPB,MTCNN,DB_VAE_algorithmic_bias}                                         \\
                              &                                                                    &                                                  & Other targets associated with protected attribute    & \cite{representation_disparity, BR_Net_dataset_vs_task,fairfil}                                                    \\
                              \midrule
\multirow{5}{*}{Type II Bias} & \multirow{5}{*}{Dependence between model prediction and attribute} & \multirow{3}{*}{Is attribute known and labeled?} & Known and labeled                             & \cite{CSAD, Back_MI, learn_not_to_learn_Colored_MNIST}                                                     \\
                              &                                                                    &                                                  & Known but unlabeled                           & \cite{HEX_texture_bias1, ReBias_texture_bias2,rubi}                                                                \\
                              &                                                                    &                                                  & Unknown                                       & \cite{ECS,LfF_CelebA_Bias_conflicting,UBNet}                                                                                    \\
                             \cmidrule{3-5} 
                              &                                                                    & \multirow{2}{*}{Type of attribute}               & Sensitive attribute/protected attribute       & \cite{machine_bias,DP_FFVAE,LAFTR}                                                                         \\
                              &                                                                    &                                                  & Spurious attribute                            & \cite{Group_DRO,End, BCL}     \\
                              \bottomrule
\end{tabular}
}
   
\end{table*}

\section{Confusion}
\label{sec:confusion}

In the previous section, we categorize \pc papers, that discuss prevalent biases, into two groups based on the manifestation of bias they address.
The criteria for this categorization are clearly outlined in~\cref{tab:unification}. 
Furthermore, the distinctions between these two types of biases are illustrated in~\cref{def:Type_I_Bias,def:Type_II_Bias}.
However, as summarized in~\cref{tab:confusion}, there is substantial confusion between them in existing literature, which poses challenges for researchers to investigate bias issues. 
Thus, it is crucial to clarify the confusion and underscore the distinctions between these two types of biases.
To this end, in this section, we primarily highlight several prevailing confusions and the potential consequences that arise from overlooking them, based on the investigation of \pc papers.
In the following sections, we analyze the possible reasons behind these confusions (\cref{sec:reason}) and provide a clear distinction between these biases to alleviate these confusions (\cref{sec:distinction}).

\begin{table}[htbp]
\caption{The summary of the existing confusion in the literature regarding bias issues.}
\label{tab:confusion}
\centering

\begin{tabular}{lc}
\toprule
Type of confusion                 & Examples                                                            \\
\midrule
Ambiguity of Terminology          & \cite{DI,BA,DB_VAE_algorithmic_bias}                                \\
Inaccurate Motivation             & \cite{Back_MI, BlindEye_IMDB_eb,FairCal}                            \\
Lack of Terminology Reuse         & \cite{minority_group_vs_sensitive_attribute,BR_Net_dataset_vs_task,spurious_correlation_Underrepresentation} \\
Abuse of Bias Assessment Metrics & \cite{FURL_PS,Fairalm_DP_difference_false_positive_rate,model_leakage}            \\
Weak Existing Distinction         & \cite{causal_based_fairness,MLbias_survey,datasets_ML}             \\
\bottomrule
\end{tabular}

\end{table}

\subsection{Ambiguity of Terminology}
One of the confusions is the ambiguity surrounding the terminology of bias.
This ambiguity manifests in three primary ways.
First, several papers adopt vague terminology such as ``bias issues" or simply ``bias" without clarifying the particular type of bias they address~\cite{DI}.
Furthermore, other commonly used terms such as ``model bias" or ``algorithmic bias" are also ambiguous, as they might represent either the bias that manifests in the model or the bias that originates from the model itself. 
Second, studies often denote bias from varied aspects~\cite{hirota2022gender, markl2022language}.
For instance, some papers refer to ``demographic bias", ``gender bias", or ``racial bias", emphasizing bias from the perspective of demographic statistics.
In contrast, other works utilize ``dataset bias", ``model bias", or ``algorithmic bias", indicating the source of bias.
Third, the existing literature frequently uses the same terms to describe different kinds of biases~\cite{MvCoM, Back_MI}, as summarized in~\cref{tab:bias_term}.

\begin{table}[htbp]
\caption{The summary of terms commonly used for bias.}
\label{tab:bias_term}
\centering
\resizebox{0.48\textwidth}{!}{%

\begin{tabular}{lccc}
\toprule
\multirow{2}{*}{Paper}             & \multirow{2}{*}{Claimed bias to address (Motivation)} & \multicolumn{2}{c}{Actual type of bias to address (Technique)} \\
\cmidrule(lr){3-4} 
                                   &                                          & Type I Bias              & Type II Bias            \\
                                   \midrule
\cite{RFW_IMAN}                         & Racial bias                              & \checkmark               &                         \\
\cite{pass}                        & Gender bias, skintone bias               & \checkmark               & \checkmark              \\
\cite{von_Mises_Fisher}            & Gender bias                              & \checkmark               & \checkmark              \\
\cite{model_leakage}            & Gender bias                              &                          & \checkmark              \\
\cite{BA}                          & Gender bias                              &                          & \checkmark              \\
\cite{DI} & Gender bias                              & \multicolumn{1}{l}{}     & \checkmark              \\
\midrule
\cite{DB_VAE_algorithmic_bias}     & Algorithmic bias                        & \checkmark               &                         \\
\cite{MvCoM}                       & Dataset bias                             & \checkmark               &                         \\
\cite{BR_Net_dataset_vs_task}             & Dataset bias                             & \checkmark               & \multicolumn{1}{l}{}    \\
\cite{Back_MI}                     & Dataset bias                             & \multicolumn{1}{l}{}     & \checkmark              \\
\cite{confused_dataset_bias_DFA}       & Dataset bias                             & \multicolumn{1}{l}{}     & \checkmark             \\
\bottomrule
\end{tabular}

}

\end{table}

\noindent
\textbf{Consequences.}
The ambiguity of terminology undermines the clarity of the intended statement and may further lead to misdirected debiasing techniques. 
For instance, in the abstract of the paper~\cite{BA}, the authors claim that:
\begin{itemize}
    \item \emph{``We find that (a) datasets for these tasks contain significant gender bias and (b) models trained on these datasets further amplify existing bias."}~\cite{BA}
\end{itemize}
In this case, the lack of clarity around the term ``gender bias" weakens the significance of the findings. 
Furthermore, the scope of this ambiguity is extensive.
Specifically, sections including ``Title", ``Abstract", ``Introduction" and ``Related Work" are often impacted, as there may lack sufficient context for a precise interpretation~\cite{sadeghi2019global, gordaliza2019obtaining}.
More concerned, the vagueness may persist throughout the entire paper~\cite{DB_VAE_algorithmic_bias} if the addressed bias is not dis-ambiguously clarified in ``Problem Statement" or evaluation protocol in ``Experiments" section.

\subsection{Inaccurate Motivation}
Another confusion is that existing work addressing these two types of bias inaccurately cites each other for their own motivation. 
For instance, some studies~\cite{Back_MI, BlindEye_IMDB_eb} that address Type II Bias motivate themselves from the uneven performance in face recognition, a manifestation of Type I Bias. 
Other work~\cite{DeepFR_survey,FairCal} that tackles Type I Bias in debiasing face recognition is motivated by the correlation between model predictions and spurious attributes in facial attribute classification~\cite{BlindEye_IMDB_eb}, a manifestation of Type II Bias. 
Furthermore, this confusion is aggravated as some papers are motivated by semi-relevant work.
Specifically, as highlighted by~\cite{FVRT3}, debiasing face recognition literature~\cite{FairCal,RFW_IMAN,RL_RBN} tend to be motivated by the manifestation of worse accuracy for minority groups in sex classification~\cite{Timnit_sex_classification_PPB}, rather than the direct issue of uneven performance in face recognition~\cite{robinson2020face, pahl2022female}.

\noindent
\textbf{Consequences.}
Inaccurate motivation leads to misunderstanding and misalignment in the existing literature. 
Furthermore, this issue may compound over time, as the subsequent work built upon the papers with such inaccurate motivation will perpetuate the confusion.

\subsection{Lack of Terminology Reuse}
The confusion also manifests in the introduction of overfull new terms in different papers addressing the same bias.
For instance, ``minority group bias"~\cite{minority_group_vs_sensitive_attribute}, ``dataset bias"~\cite{BR_Net_dataset_vs_task}, and ``bias as underrepresentation"~\cite{spurious_correlation_Underrepresentation} are all used to denote uneven performance across attributes (Type I Bias).
\begin{itemize}
    \item \emph{``Dataset bias is often introduced due to the lack of enough data points spanning the whole spectrum of variations with respect to one or a set of protected variables."}~\cite{BR_Net_dataset_vs_task}
    \item \emph{``Minority group bias. When a subgroup of the data has a particular attribute or combination of attributes that are relatively uncommon compared to the rest of the dataset, they form a minority group. A model is less likely to correctly predict for samples from a minority group than for those of the majority."}~\cite{minority_group_vs_sensitive_attribute}
    \item \emph{``[...] `bias' means that one appearance of an object is underrepresented."}~\cite{spurious_correlation_Underrepresentation}
\end{itemize}

\noindent
Similarly, ``sensitive attribute bias"~\cite{minority_group_vs_sensitive_attribute}, ``task bias"~\cite{BR_Net_dataset_vs_task}, and ``bias as spurious correlation"~\cite{spurious_correlation_Underrepresentation} all signify the dependence between model prediction and attribute (Type II Bias).
\begin{itemize}
    \item \emph{``Task bias, on the other hand, is introduced by the intrinsic dependency between protected variables and the task."}~\cite{BR_Net_dataset_vs_task}
    \item \emph{``Sensitive attribute bias. A sensitive attribute (also referred to as ``protected") is one which should not be used by the model to perform the target task, but which provides an unwanted “shortcut” which is easily learned, and results in an unfair model."}~\cite{minority_group_vs_sensitive_attribute}
    \item \emph{``[...] considering bias in the form of spurious correlations between the target label and a sensitive attribute which is predictive on the training set but not necessarily so on the test set."}~\cite{spurious_correlation_Underrepresentation} 
\end{itemize}

\noindent
\textbf{Consequences.}
These inconsistent definitions can further contribute to confusion with some highlighting the manifestation of the bias while others delving into the underlying causes of the bias.
Furthermore, without a unified terminology for the predominant biases, it becomes challenging to systematically gather and compare relevant work.

\subsection{Abuse of Bias Assessment Metrics}
The usage of bias assessment metrics exhibits the confusion in two primary ways.
First, the bias assessment metrics, which are designed independently of debiasing methods, are rarely used~\cite{RLB,Directional_BA}.
Instead, many works tend to introduce their own metrics to demonstrate the effectiveness of the proposed debiasing method~\cite{model_leakage,BA}, which leads to an overwhelming number of metrics.
Second, some studies inappropriately employ indirect bias assessment metrics or even metrics that are not designed for the specific bias they address.
For instance, several studies~\cite{FURL_PS,Fairalm_DP_difference_false_positive_rate} motivated by the dependence between model prediction and attributes (the manifestation of Type II Bias) use true positive rate (TPR) difference and false positive rate (FPR) difference for evaluation. 
However, as highlighted by~\cite{Directional_BA}, metrics such as TPR difference, FPR difference, accuracy difference, and average mean-per-class accuracy difference, are not suitable for evaluating Type II Bias since they fail to consider the dependence between target and attribute in the training set and cannot distinguish between an increase or decrease of dependence in learned representation. 

\noindent
\textbf{Consequences.}
The abuse of bias assessment metrics leads to inaccurate evaluations of debiasing performance in relation to the specific type of bias being addressed, hence exacerbating confusion in the field.
Furthermore, it also complicates the comparison between different debiasing methods and hinders the construction of a unified evaluation protocol.

\subsection{Weak Existing Distinction}
Despite the evident confusion in the literature, numerous studies, especially survey papers, have not sufficiently distinguished Type I Bias and Type II Bias. 
Furthermore, the confusion is not only widespread but has also persisted for a significant duration, as shown by the timeframes of the investigated papers.
However, the bias taxonomy, presented in surveys over time~\cite{causal_based_fairness,MLbias_survey,datasets_ML}, may fail to clearly differentiate between these two types of biases. 
Alarmingly, a recent and high-cited survey on machine learning bias~\cite{MLbias_survey} scarcely cites papers that discuss Type II Bias stemming from spurious correlations between target and attribute, thereby overlooking the distinction from Type I Bias.

\noindent
\textbf{Consequences.} 
The weak distinction between these two types of biases in existing surveys will exacerbate the prevailing confusion in this field over time. 
Consequently, due to the lack of clarity, which surveys were originally designed to provide concerning the categorization of bias issues, these bias issues will eventually be undesirably conflated.

\section{Reasons of Confusion}
\label{sec:reason}
In this section, we investigate various factors that may contribute to the confusion discussed in the previous section. 
Specifically, we examine the historical context, the preconception about bias, and the methodologies adopted to address different biases, to provide insights on how and why such confusion has persisted in the literature.

\begin{figure*}[t]
\centering
  \includegraphics[width=0.7\linewidth]{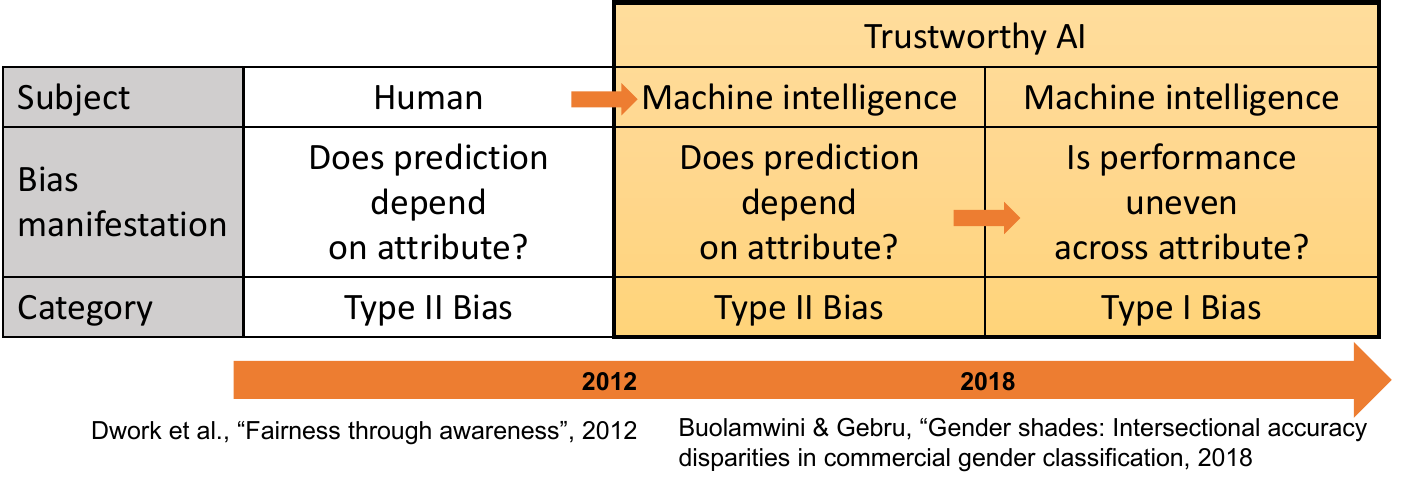}
  \caption{
  The enrichment of the concept ``bias" in machine intelligence with important milestones. Initially, ``bias" implied that human decision-making depends on protected attributes (Type II Bias). As machine intelligence began aiding human decision-making processes, the subject of ``bias" broadened from humans to algorithms. Along with the continued advances of machine intelligence, a new aspect of bias issues, performance disparity across demographic groups (Type I Bias), further enriched the meaning of ``bias".
  Currently, addressing both Type I Bias and Type II Bias becomes essential for ensuring Trustworthy AI.
  }
\label{fig:enrichment}
\end{figure*}

\subsection{Historical Context}
We first examine the historical origins of bias issues.
In~\cref{fig:enrichment}, we summarize the enrichment of the concept ``bias" in machine learning from the perspective of Type I Bias and Type II Bias and highlight key milestones throughout its history.
Originally, ``bias" is defined as unfair favoritism or prejudice towards one thing, person, or group over another~\cite{ditomaso2015racism}.
Specifically, bias issues are especially evident in real-world decision-making processes, such as advertising, financial creditworthiness, employment, education, and criminal justice~\cite{ruggeri2023persistence,edmond2019just}.
To promote fairness, certain sensitive attributes (\eg sex, age, and race) are by law defined as protected attributes that cannot be discriminated against in the decision-making process~\cite{protected_attributes}.
In this initial stage, decisions are primarily made by humans.
Thus, the main bias issue is if human decision-making depends on protected attributes, which aligns with Type II Bias in our definitions.

Following the emergence of neural networks, machine learning models start to assist in human decision-making processes~\cite{bastani2021improving,dankwa2019transforming}. 
This evolution also leads to an expansion of the subject in the discussion regarding bias issues, from human decision-making to algorithmic decision-making~\cite{starke2022fairness}. 
With this change, numerous works begin to explore if algorithmic decision-making depends on protected attributes (\ie demographic parity)~\cite{fairness_through_awareness, counterfactual_fairness}, which also align with Type II Bias.
Meanwhile, along with the advancement of neural networks, its performance becomes a crucial evaluation criterion.
Consequently, it brings significant attention to a new aspect of bias issues: performance disparity across demographic groups~\cite{Timnit_sex_classification_PPB,Accuracy_parity}, which aligns with Type I Bias in our definitions.
Furthermore, new fairness criteria such as equalized odds and equal opportunity~\cite{EO_define}, which address disparities in true positive rates and false positive rates across demographic groups, are adopted from demographic parity.

We conjecture that the confusion arises because the term ``bias" in neural networks has been endowed with multiple important meanings over time without well-defined distinctions. This ambiguity leads individuals to interpret different types of predominant biases from the same term.
Specifically, some individuals associate the primary bias with performance disparity due to the critical role of model performance in model evaluation.
Conversely, other individuals prioritize prediction disparity since it is the prevalent bias deeply embedded in real-world scenarios.
Consequently, denoting these two different but predominant biases with the single term ``bias" results in misunderstandings in the broader literature.

\subsection{Preconception about Bias}
The preconception of researchers about bias, stemming from their specific relevant fields, also contributes to the confusion.
Specifically, bias issues encompass a wide range of relevant fields, some of which are associated with Type I Bias and others with Type II Bias.
For instance, Type I Bias involves long-tail distribution~\cite{long_tail}, catastrophic forgetting~\cite{kirkpatrick2017overcoming}, domain adaptation~\cite{li2014learning}, and various biometric tasks~\cite{xiao2023name,hutiri2022bias}.
In contrast, Type II Bias involves shortcut learning~\cite{shortcut_learning}, simplicity bias~\cite{simplicitybias}, invariant representation learning~\cite{DP_FFVAE}, out-of-distribution challenges~\cite{shen2021towards}.
In this sense, researchers from diverse fields hold their own preconceived notions of bias based on their field-specific knowledge. 
For instance, in several biometric tasks (\eg face recognition, face 
detection, face verification) with identity as target and sex as an attribute, uneven performance across sex (the manifestation of Type I Bias) is naturally regarded as bias since the primary focus of biometric systems is on model performance~\cite{robinson2020face}.
However, the dependence between model prediction and attribute (the manifestation of Type II Bias) might not be considered as bias since there naturally only exists non-overlapping targets across attribute~\cite{RFW_IMAN}. For instance, an individual can be categorized as either male or female but not both, thereby resulting in a natural association between identity prediction and specific sex.
Furthermore, due to the absence of clear distinctions regarding bias issues, research groups from different fields may not share a unified perspective on bias and may interpret it differently. However, they use similar bias-related terms in their papers and present them in the same venues, which potentially causes confusion regarding bias issues.

\subsection{Similar Methodologies}
The existing confusion also arises from the overlap in methodologies used to address Type I Bias and Type II Bias.
For instance, to mitigate Type I Bias, several studies~\cite{SensitiveNets,DebFace,pass} enhance the performance for minority groups by preventing the model from encoding the information of protected attribute.
Similarly, to tackle Type II Bias, some methods~\cite{Back_MI,CSAD,learn_not_to_learn_Colored_MNIST} aim to develop representations that are invariant to the protected attribute by minimizing mutual information between the learned representation and the protected attribute.
Both of these methods can be categorized into invariant representation learning~\cite{IRM}.
Furthermore, domain adaptation is also utilized for both Type I Bias~\cite{BAE,MFR} and Type II Bias~\cite{DARE}.
These similarities in methodologies obscure the distinction between Type I Bias and Type II Bias, thereby inducing confusion.

\section{Experimental Discussion}
\label{sec:distinction}

\begin{figure*}[htbp]
    \begin{minipage}{0.65\textwidth}

      \subfloat[Training set.]{\includegraphics[width=0.38\linewidth]{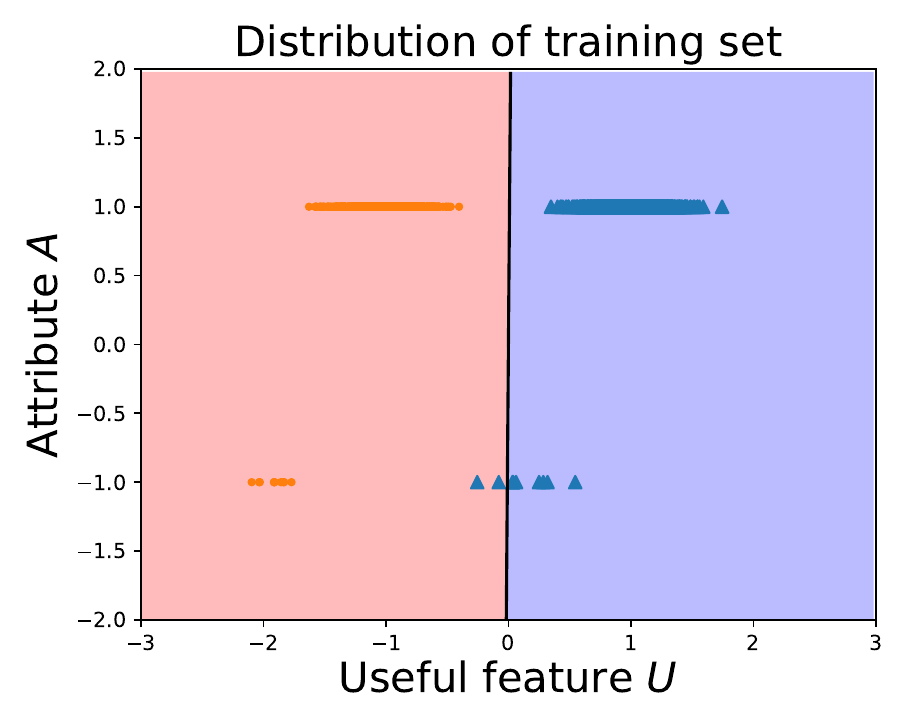}\label{fig:I_exists_train}}\quad
      \subfloat[Testing set.]{\includegraphics[width=0.62\linewidth]{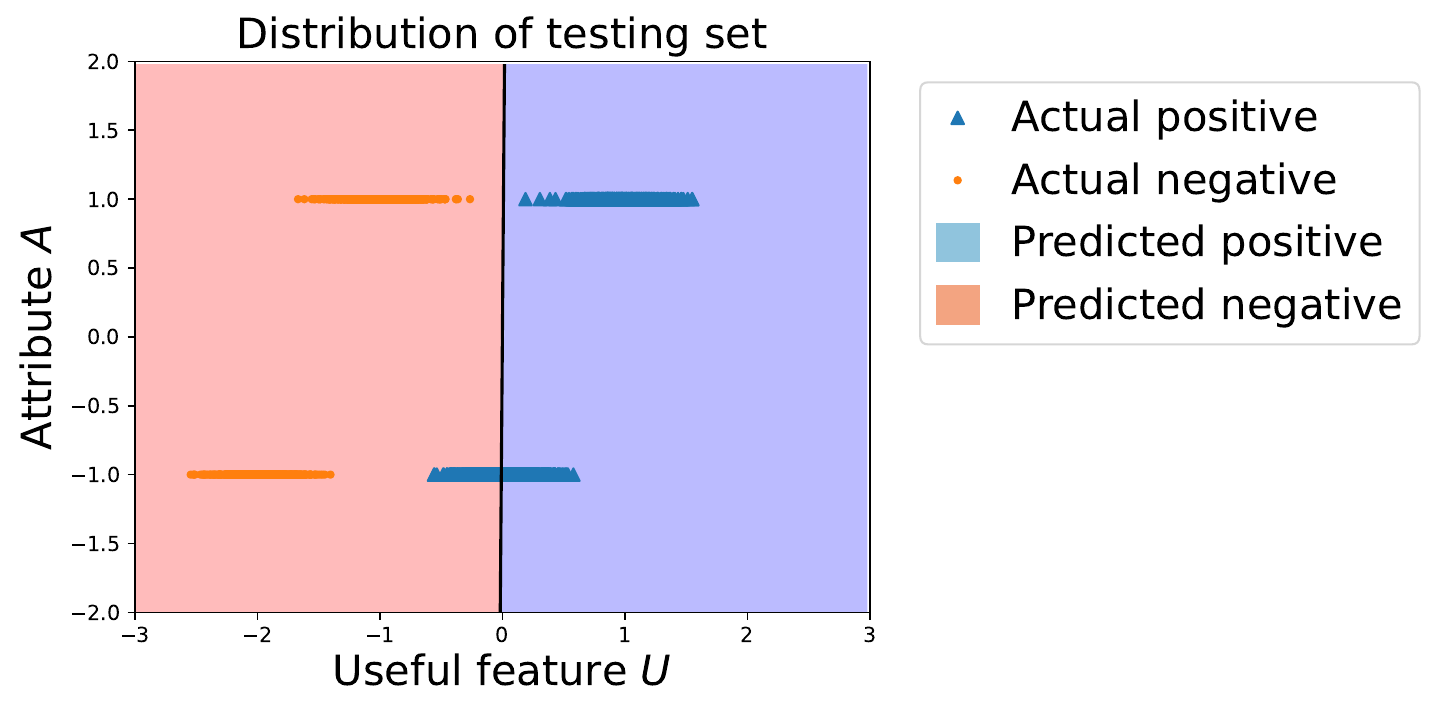}\label{fig:I_exists_test}}\quad
        \caption{Distribution of training and testing sets regarding synthetic data. The vertical classification boundary (labeled as the black line) reveals that the classifier does not utilize $A$ for classification. However, there are more wrong predictions in the group of $A=-1$ than in the group of $A=1$, which violates performance parity.}
        \label{fig:I_exists}

    \end{minipage}%
    \hspace{0.3cm}
    \begin{minipage}{0.35\textwidth}
        \captionsetup{type=table}
        \caption{Type I Bias exists without Type II Bias since there exists accuracy disparity across $A$ while $\hat{Y}$ and $A$ are independent.}
        \label{tab:I_exists}
        \centering
        \resizebox{1\textwidth}{!}{%

\begin{tabular}{lccc}
\toprule
           & Accuracy & $P(\hat{Y}=0|A)$ & $P(\hat{Y}=1|A)$ \\
           \midrule
$A=1$      & 100.00   & 66.7\%           & 33.3\%           \\
$A=-1$     & 65.33    & 66.7\%           & 33.3\%           \\
$|\Delta|$ & 34.67    & 0                & 0               \\
\bottomrule
\end{tabular}
        
        }
    \end{minipage}
\end{figure*}

In this section, we empirically investigate the distinction between Type I Bias and Type II Bias.
Specifically, we conduct experiments on two synthetic datasets and two well-known real-world datasets: Adult Income Dataset~\cite{adult_dataset_and_german_dataset} and CelebA Dataset~\cite{CelebA}.
First, we use synthetic data to demonstrate that Type I Bias and Type II Bias are unrelated, \ie one can exist without the presence of the other bias.
Next, we utilize Adult dataset to further illustrate the difference between Type I Bias and Type II Bias in real-world scenarios.
Last, we employ CelebA dataset to evaluate the effectiveness of multiple representative bias assessment metrics in assessing Type I Bias and Type II Bias.
All experimental results are obtained by averaging the results over 10 trials.

\subsection{Unrelated Occurrence}
In this section, we leverage synthetic data to simulate two scenarios: the first scenario showcases the presence of Type I Bias without Type II Bias, while the second scenario showcases the presence of Type II Bias without Type I Bias.

\noindent
\textbf{Setup.}
We construct the synthetic dataset containing instances $(x, y)$, where $x$ denotes a two-dimensional input consisting of the useful feature $u$ and the binary attribute $a$, and $y$ denotes the target label.
Next, we apply a classifier $C: \mathcal{X} \rightarrow \mathcal{Y}$ to consume the input $x$ and produce the prediction $\hat{y} = C(x) = C(u,a) \in \mathcal{Y}$. 
The classifier is a single fully connected layer (FC) followed by the binary cross-entropy loss.
To evaluate Type I Bias, we measure the difference in accuracy. To assess Type II Bias, we utilize the Calders-Verwer discrimination score~\cite{discrimination_score} defined as $|P(\hat{Y}=y|A=1) - P(\hat{Y}=y|A=-1)|$.

\begin{figure*}[htbp]
    \begin{minipage}{0.65\textwidth}

      \subfloat[Training set.]{\includegraphics[width=0.38\linewidth]{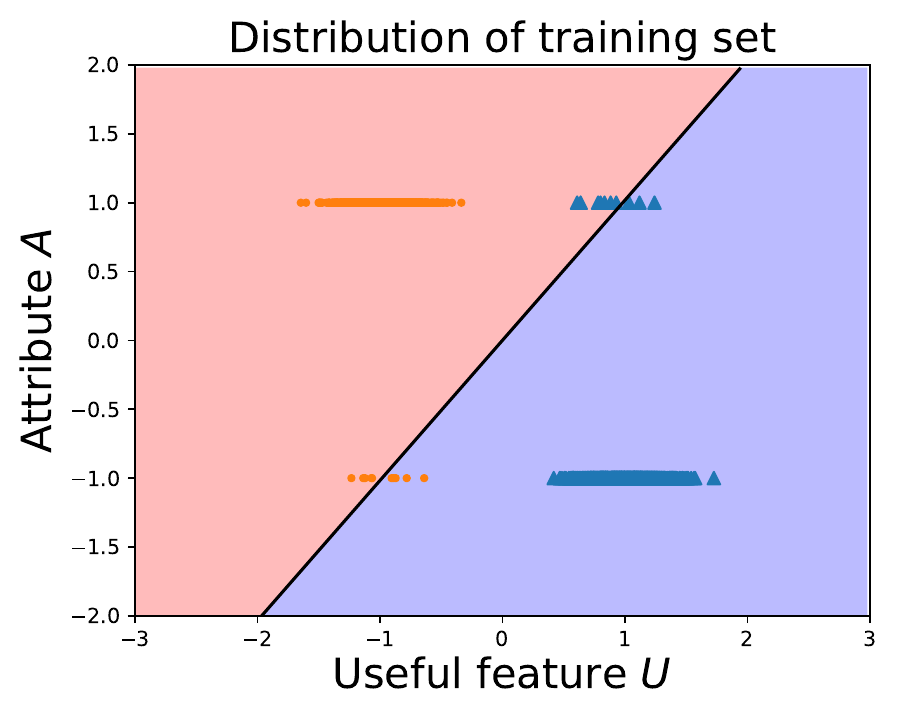}\label{fig:II_exists_train}}\quad
      \subfloat[Testing set.]{\includegraphics[width=0.62\linewidth]{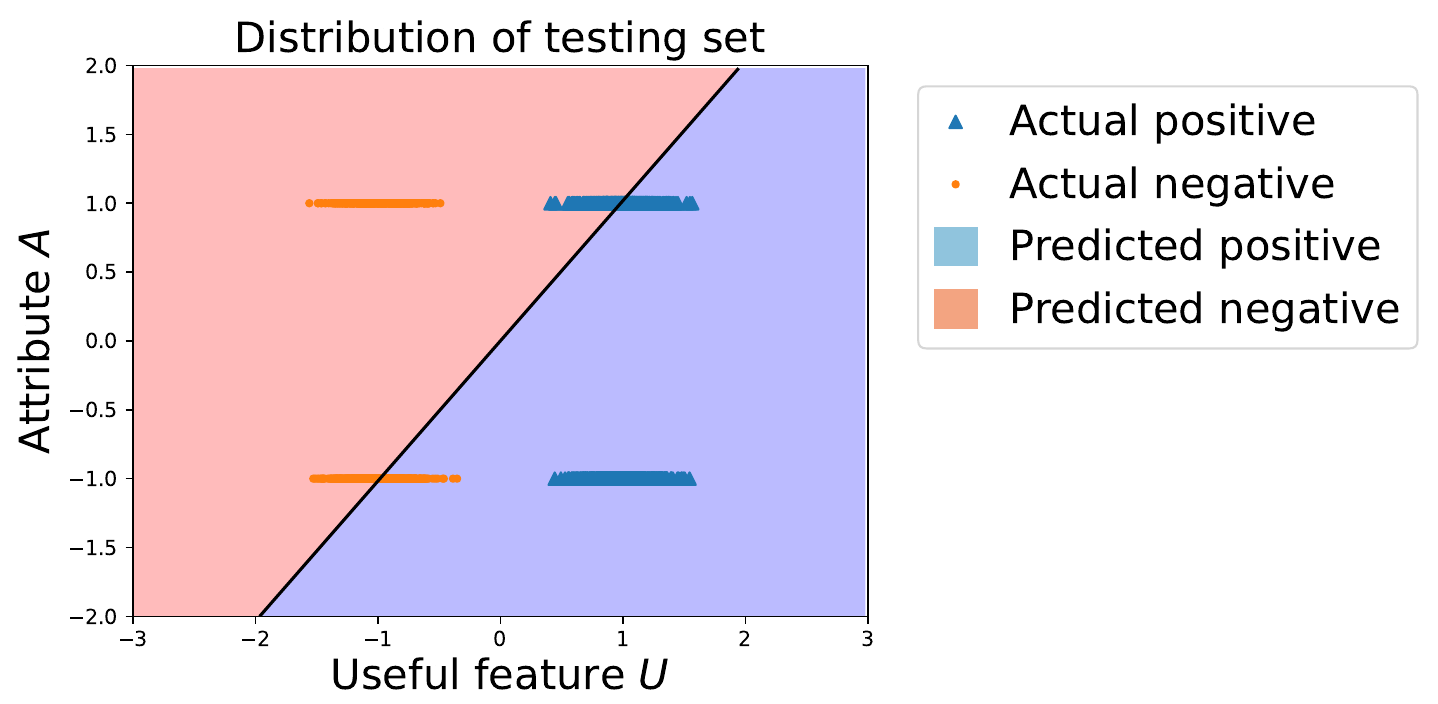}\label{fig:II_exists_test}}\quad
      \caption{Distribution of training and testing sets regarding synthetic data. The non-vertical classification boundary (labeled as the black line) reveals that the classifier utilizes $A$ for classification. However, the number of wrong predictions is approximately the same across $A$, thereby fulfilling performance parity.}
        \label{fig:II_exists}
        
    \end{minipage}%
    \hspace{0.3cm}
    \begin{minipage}{0.35\textwidth}
        \captionsetup{type=table}
        \caption{Type II Bias exists since $\hat{Y}$ and $A$ are not independent while there is no accuracy disparity across $A$.}
        \label{tab:II_exists}
        \centering
        \resizebox{1\textwidth}{!}{%

\begin{tabular}{lccc}
\toprule
           & Accuracy & $P(\hat{Y}=0|A)$ & $P(\hat{Y}=1|A)$ \\
           \midrule
$A=1$      & 85.98    & 64.1\%           & 35.9\%           \\
$A=-1$     & 85.97    & 35.4\%           & 64.6\%           \\
$|\Delta|$ & $\approx 0$     & 28.7\%           & 28.7\%          \\
\bottomrule
\end{tabular}
        
        }%
    \end{minipage}
\end{figure*}

\subsubsection{Type I Bias exists without Type II Bias}
We synthesize training set \wrt $A,X,Y$ by the following generative model,

\begin{align*}
&A \sim \text{Ber}(1/100) \times 2 - 1; \\
&V_1 \sim \text{Norm}(-1,\sigma=0.2); \\
&V_2 \sim \text{Norm}(1,\sigma=0.2); \\
&T \sim \text{Ber}(1/2); \\
&U|_{A=1} \sim V_1 \times T + V_2 \times (1-T); \\
&U|_{A=-1} \sim U|_{A=1} - 1; \\
&X = [U,A]^T; \\
&Y \sim \mathbbm{1}_{U > 0};
\end{align*}

\noindent
where $\text{Ber}(p)$ represents the Bernoulli distribution with probability $p$, $\text{Norm}(\mu, \sigma)$ represents the normal distribution with mean $\mu$ and standard deviation $\sigma$, and $\mathbbm{1}$ is the indicator function. 
As shown in~\cref{fig:I_exists}, the training set is imbalanced across attribute $A$, with the subset where $A=-1$ being the minority group. 
Furthermore, the optimal classification boundary is set to be varied across $A$ since one widely accepted cause of Type I Bias is that the model trained on the sufficient samples in majority groups might not effectively generalize to minority groups~\cite{spurious_correlation_Underrepresentation}. 
Additionally, the testing set is constructed using the following generative model,
\begin{align*}
&A \sim \text{Ber}(1/2) \times 2 - 1; \\
&V_1 \sim \text{Norm}(-1,\sigma=0.2); \\
&V_2 \sim \text{Norm}(1,\sigma=0.2); \\
&T \sim \text{Ber}(1/3); \\
&U|_{A=1} \sim V_1 \times T + V_2 \times (1-T); \\
&U|_{A=-1} \sim V_1 \times T + V_2 \times (1-T) - 1; \\
&X = [U,A]^T; \\
&Y \sim \mathbbm{1}_{X > 0};
\end{align*}

\noindent
where $A$ is assigned either value 0 or 1 with equal probability.
Hence, the testing set is balanced across values of the attribute.

\noindent
\textbf{Analysis.}
In~\cref{fig:I_exists}, we observe that the learned classification boundary is vertical at $X=0$, which is primarily determined by dominant samples in the majority group.
The vertical boundary suggests that the model does not use attribute $A$ for classification.
Furthermore, as highlighted in~\cref{tab:I_exists}, given that $P(\hat{Y}=y|A=1) = P(\hat{Y}=y|A=-1)~\forall~y \in \{0,1\}$, model prediction $\hat{Y}$ is independent with attribute $A$, \ie Type II Bias does not exist.
However, it is noteworthy that there is a significant performance disparity between the majority and minority groups, which confirms the existence of Type I Bias.

\subsubsection{Type II Bias exists without Type I Bias}
We synthesize training set \wrt $A,X,Y$ by the following generative model,
\begin{align*}
&A \sim \text{Ber}(1/2) \times 2 - 1; \\
&V_1 \sim \text{Norm}(-1,\sigma=0.2); \\
&V_2 \sim \text{Norm}(1,\sigma=0.2); \\
&T \sim \text{Ber}(1/100); \\
&U|_{A=1} \sim V_1 \times (1-T) + V_2 \times T; \\
&U|_{A=-1} \sim V_1 \times T + V_2 \times (1-T); \\
&X = [U,A]^T; \\
&Y \sim \mathbbm{1}_{X > 0}.
\end{align*}

\noindent
As shown in~\cref{fig:II_exists}, the training set yields more samples with combinations $A=1,Y=0$ and $A=-1,Y=1$ compared to other combinations.
This setting is motivated by that the association between target $Y$ and attribute $A$ in the training set is considered one widely-accepted reason for Type II Bias~\cite{LfF_CelebA_Bias_conflicting,CSAD,End}.
The testing set is generated to be balanced across both $Y$ and $A$ with the following generative model,

\begin{align*}
&A \sim \text{Ber}(1/2) \times 2 - 1; \\
&V_1 \sim \text{Norm}(-1,\sigma=0.2); \\
&V_2 \sim \text{Norm}(1,\sigma=0.2); \\
&T \sim \text{Ber}(1/2); \\
&U|_{A=1} \sim V_1 \times (1-T) + V_2 \times T; \\
&U|_{A=-1} \sim V_1 \times T + V_2 \times (1-T); \\
&X = [U,A]^T; \\
&Y \sim \mathbbm{1}_{X > 0}.
\end{align*}

\noindent
\textbf{Analysis.}
In~\cref{fig:II_exists}, we observe that the learned classification boundary is not vertical, which suggests that the classifier relies on $A$ for decision-making.
Furthermore, as highlighted in~\cref{tab:II_exists}, given that $P(\hat{Y}=y|A=1) \neq P(\hat{Y}=y|A=-1)~\forall~y \in \{0,1\}$, model prediction $\hat{Y}$ is not independent with attribute $A$, \ie Type II Bias exists.
However, for Type I Bias, it is noteworthy that there is no significant performance disparity between the majority and minority groups.

\begin{figure*}[htbp]
    \centering %

      \subfloat[Female is the minority group.]{\includegraphics[width=0.47\linewidth]{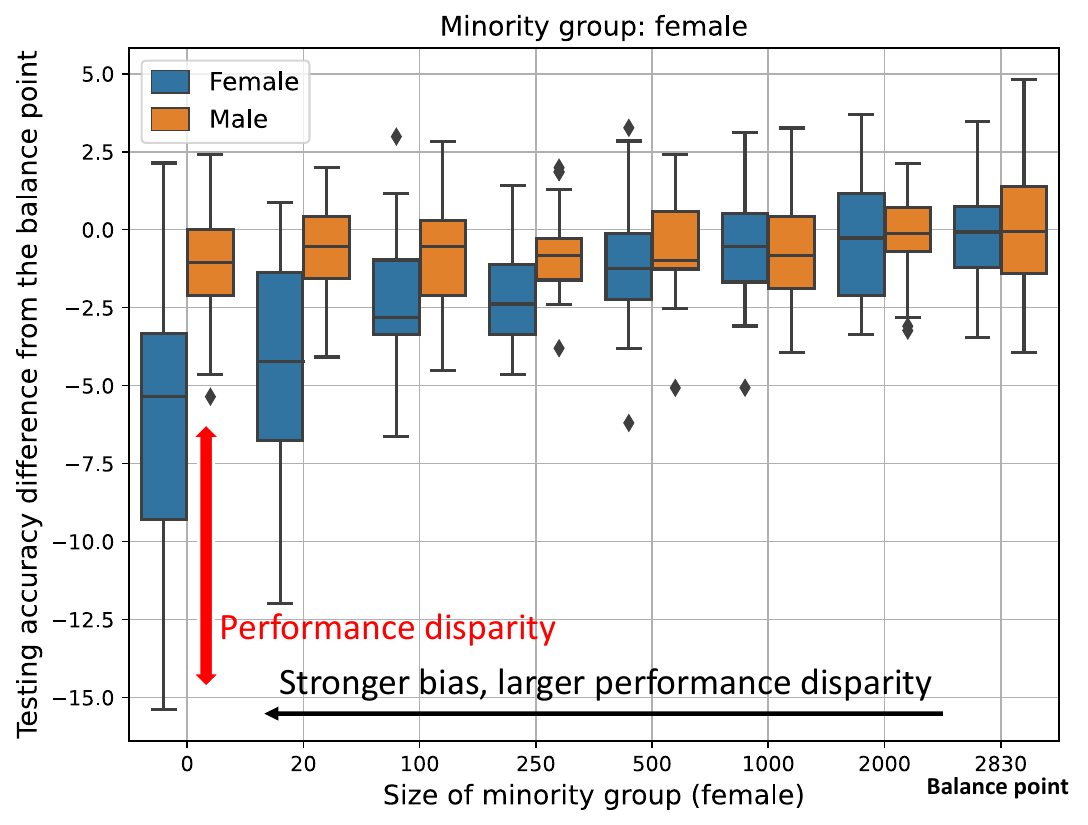}\label{fig:Adult_I_female}}\quad
      \subfloat[Male is the minority group.]{\includegraphics[width=0.46\linewidth]{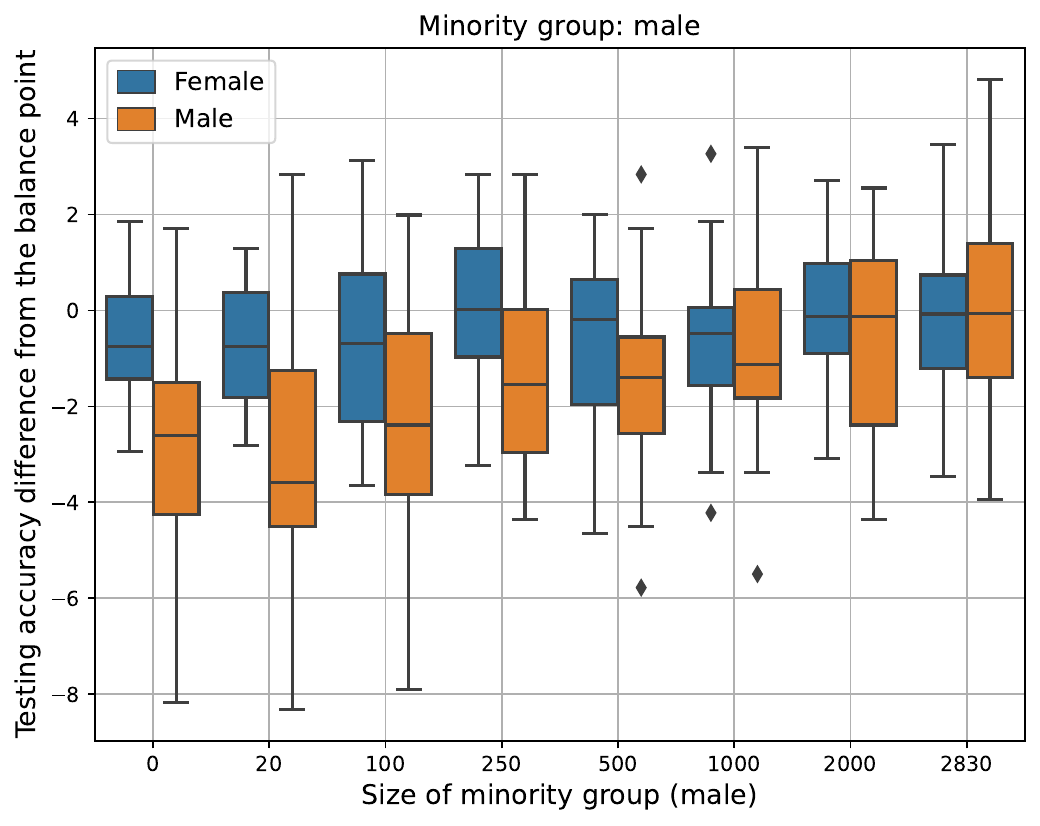}\label{fig:Adult_I_male}}\quad

\caption{Illustration of Type I Bias on Adult which manifests as uneven performance between the minority and majority groups. As Type I Bias becomes stronger (the minority size decreases), the accuracy for minority group diminishes while the accuracy for majority group remains unchanged, thereby enlarging the performance disparity across the minority and majority groups.}
\label{fig:Adult_I}
\end{figure*}

\subsection{Different Manifestations in Real World}
In this section, we utilize Adult Income Dataset~\cite{adult_dataset_and_german_dataset} to illustrate different manifestations of Type I Bias and Type II Bias in real-world scenarios. 
Adult Dataset is a census dataset where the target is whether a person earns a higher income (over 50K USD per year) and the protected attribute is sex.
As shown in~\cref{tab:stats}, the dataset is partitioned into four quarters based on the combination of target labels and protected attribute labels, given that both are binary in nature.
The statistics illustrate that Adult dataset is well-suited for investigating both Type I Bias and Type II Bias. 
Specifically, the dataset exhibits an uneven distribution across sex, with a larger number of female individuals (16,192) compared to male individuals (32,650), which could induce Type I Bias.
Furthermore, the dataset also exhibits a substantial disparity in the number of samples with higher income between females (1,769) and males (9,918), which could induce Type II Bias.

\noindent
\textbf{Setup.}
We perform data pre-processing on input census data. 
Specifically, we transform the categorical features using one-hot encoding and normalize the numerical features into Gaussian distribution with zero mean and unit variance.
Consequently, each input sample is transformed into a 108-dimensional vector.
For the training model, we employ a three-layer multilayer perceptron (MLP) followed by the binary cross-entropy loss as the baseline classifier.

\begin{table}[htbp]
\caption{Statistics of Adult dataset. The number of females is greater than the number of males, which could induce Type I Bias. Furthermore, the number of samples with higher income and samples with lower income are different across sex categories, which could induce Type II Bias.}
\label{tab:stats}
\centering
\begin{tabular}{lccc}
\toprule
       & Higher income & Lower income & Total \\
       \midrule
Female & 1769          & 14423        & 16192 \\
Male   & 9918          & 22732        & 32650 \\
Total  & 11687         & 37155        & 48842 \\
\bottomrule
\end{tabular}
\end{table}

\subsubsection{Type I Bias}
To investigate Type I Bias, we construct several imbalanced training sets and control the bias strength by modifying the degree of imbalance in the training set.
Specifically, we initially construct a balanced training set across both target $Y$ and attribute $A$ using 80\% of the entire dataset and a balanced testing set with the remaining samples.
We then manually adjust the size of the minority group in the training set while maintaining the size of the majority group to control bias strength.
Additionally, we construct two distinct groups of training sets, with either females or males as the minority group.
For instance, considering the setting where the female is minority group and the minority size is 100, the training set would consist of 50 higher-income females and 50 lower-income females, in addition to all males from the balanced training set.
We conduct experiments under different minority sizes and present the testing performance versus the size of the minority group in~\cref{fig:Adult_I}.

\noindent
\textbf{Analysis.}
Notably, we notice a non-zero accuracy disparity between females (85.15\%{\scriptsize $\pm$1.52}) and males (78.38\%{\scriptsize $\pm$1.90}) at the balance point where the training set is evenly distributed across both target $Y$ and attribute $A$.
We conjecture that this disparity is mainly because certain groups are inherently more difficult to classify than other groups~\cite{FR_inherent_bias}.
To facilitate a clearer analysis of Type I Bias, we use the accuracy difference from the testing accuracy at the balance point to represent the testing performance. 
This difference in testing accuracy, denoted as $Acc_{\text{diff}}$, is calculated by subtracting the testing accuracy at the balance point from the absolute accuracy at a given bias strength, \ie $Acc_{\text{diff}} = Acc_{\text{abs}} - Acc_{\text{balance}}$.
In~\cref{fig:Adult_I}, we observe that the performance disparity exists across the minority group and the majority group.
The accuracy for the minority group tends to decrease as its size diminishes (bias strength increases), especially when there are very limited samples from the minority group.
Furthermore, in~\cref{fig:Adult_I_female}, we observe that stronger bias results in larger performance fluctuations (bigger spread in the boxplot), which highlights the lack of robustness under such conditions.
In summary, the manifestation of Type I Bias in real-world scenarios is uneven performance across demographic groups.
One plausible cause is the imbalance in data representation across these groups in the training set.
For instance, some demographic groups may be underrepresented due to long-tail distribution~\cite{long_tail}, resulting in a skewed distribution of samples across different demographic groups.
Consequently, while data-driven models are more accurately trained on demographic groups with sufficient samples, they may not be as effective for underrepresented groups, which leads to poor prediction accuracy and unfairness towards these groups.

\begin{figure*}[htbp]
    \centering %
    
      \subfloat[Trained on EB1 Balanced consisting of females with higher income and males with lower income.]{\includegraphics[width=0.47\linewidth]{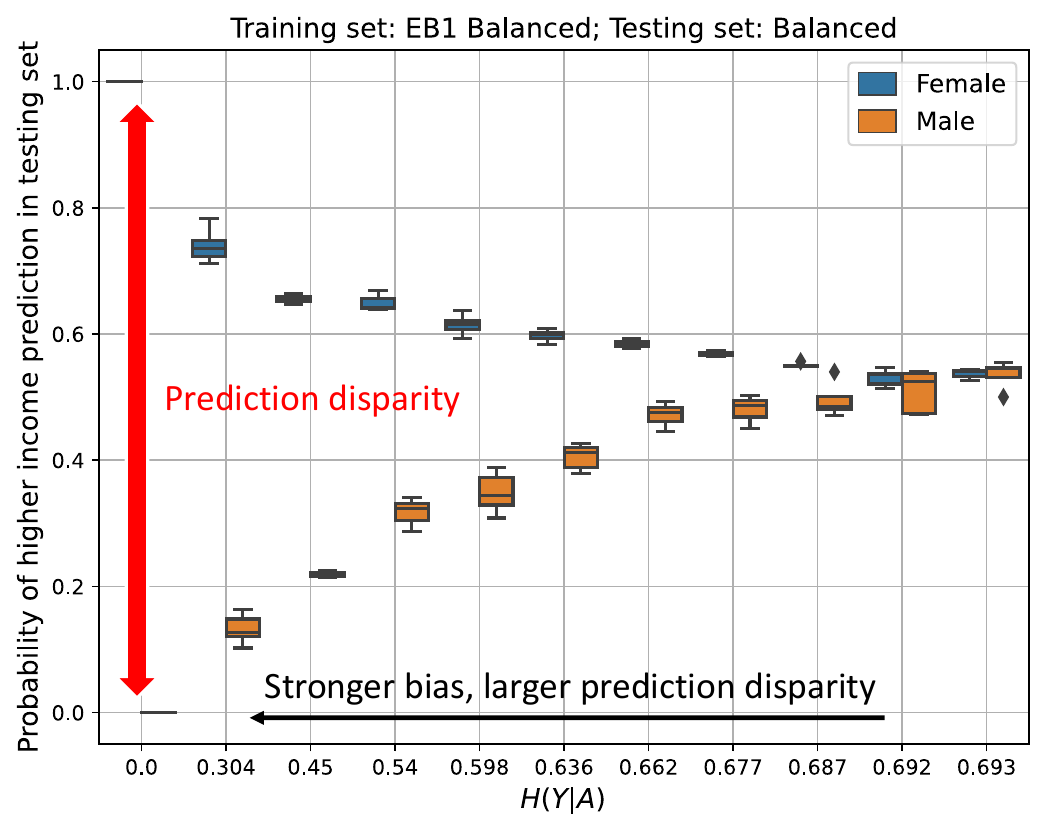}\label{fig:Adult_II_EB1_balanced}}\quad
      \subfloat[Trained on EB2 Balanced consisting of females with lower income and males with higher income.]{\includegraphics[width=0.47\linewidth]{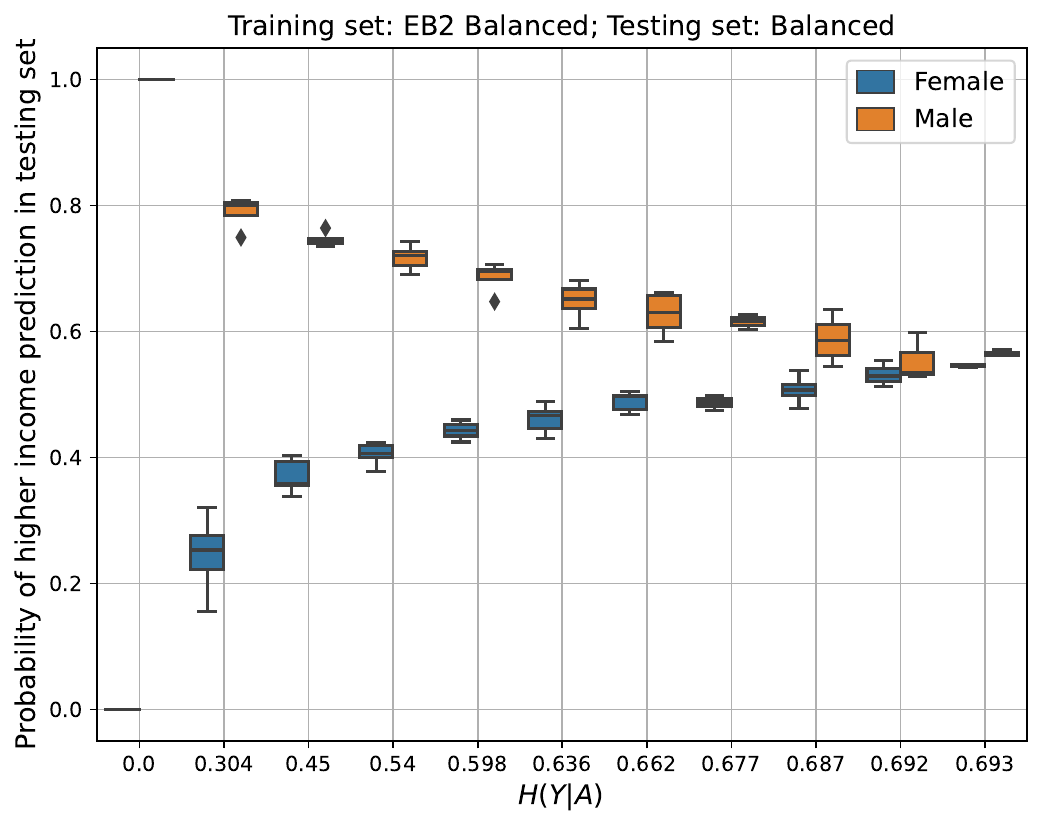}\label{fig:Adult_II_EB2_balanced}}\quad

\caption{Illustration of Type II Bias on Adult which manifests as the dependence between model prediction and attribute. 
As Type II Bias intensifies ($H(Y|A)$ decreases, rendering the attribute more predictable of the target), the prediction probability in outputting a specific prediction diverges between females and males, \ie decision-making increasingly relies on the attribute.}

\label{fig:Adult_II}
\end{figure*}

\subsubsection{Type II Bias}
To investigate Type II Bias, we construct the training set where the target $Y$ is associated with the attribute $A$ and control the bias strength by adjusting the strength of the association between $Y$ and $A$ in the training set.
Specifically, we initially construct two balanced training datasets consisting of 3538 records, each associating either females or males with higher income: (1) Extreme Bias 1 Balanced (EB1 Balanced) only contains females with higher income and males with lower income, and (2) Extreme Bias 2 Balanced (EB2 Balanced) only contains males with higher income and females with lower income.
Subsequently, we adjust the percentage of bias-conflicting samples (samples with the opposite bias present in the training set) while ensuring a consistent number of biased samples. 
This strategy enables us to construct multiple training sets, each with a distinct conditional entropy $H(Y|A)$ (\ie the smaller $H(Y|A)$, the more predictive the attribute $A$ is of the target $Y$, and the stronger the bias). 
Additionally, we construct a balanced testing set (Balanced) consisting of 7076 records ensuring an even distribution of all combinations of target and attribute labels.
Note that all these datasets are designed to be balanced across attribute to mitigate the effect of Type I Bias.

\noindent
\textbf{Analysis.}
In~\cref{fig:Adult_II}, we observe that there is a significant prediction disparity between females and males.
Furthermore, this disparity becomes more pronounced as $H(Y|A)$ diminishes (the bias strength increases).
In summary, the manifestation of Type II Bias in real-world scenarios is the dependence on the attribute in decision-making processes.
One widely accepted reason is an uneven distribution of \emph{specific target groups} across attributes, distinguishing it from Type I Bias, which emerges from an uneven distribution of samples across attributes. 
For instance, the collected dataset may contain more negative samples for female individuals and positive samples for male individuals compared to other target-attribute combinations.
During training, the model may leverage sex as the shortcut feature to simplify the learning process, rather than learning more comprehensive features. 
However, such an association between specific targets and attributes does not generally exist in the real world. 
Consequently, during applying, the trained model may still rely on the attribute, which leads to a higher frequency of positive outcomes for specific individuals and further unfair treatment for these groups.

\subsubsection{Summary}
As shown in~\cref{fig:Adult_I}, Type I Bias manifests as the performance disparity across $A$, which is evaluated based on the joint distribution of model prediction $\hat{Y}$ and ground truth $Y$.
Conversely, as shown in~\cref{fig:Adult_II}, Type II Bias manifests as the prediction disparity across $A$, which is evaluated solely based on the distribution of model prediction $\hat{Y}$.
Thus, Type I Bias and Type II Bias are unrelated phenomena and exhibit different impacts on the fairness of neural networks.

\begin{figure*}[htbp]
    \centering %
    
      \subfloat[Testing accuracy.]{\includegraphics[width=0.4\linewidth]{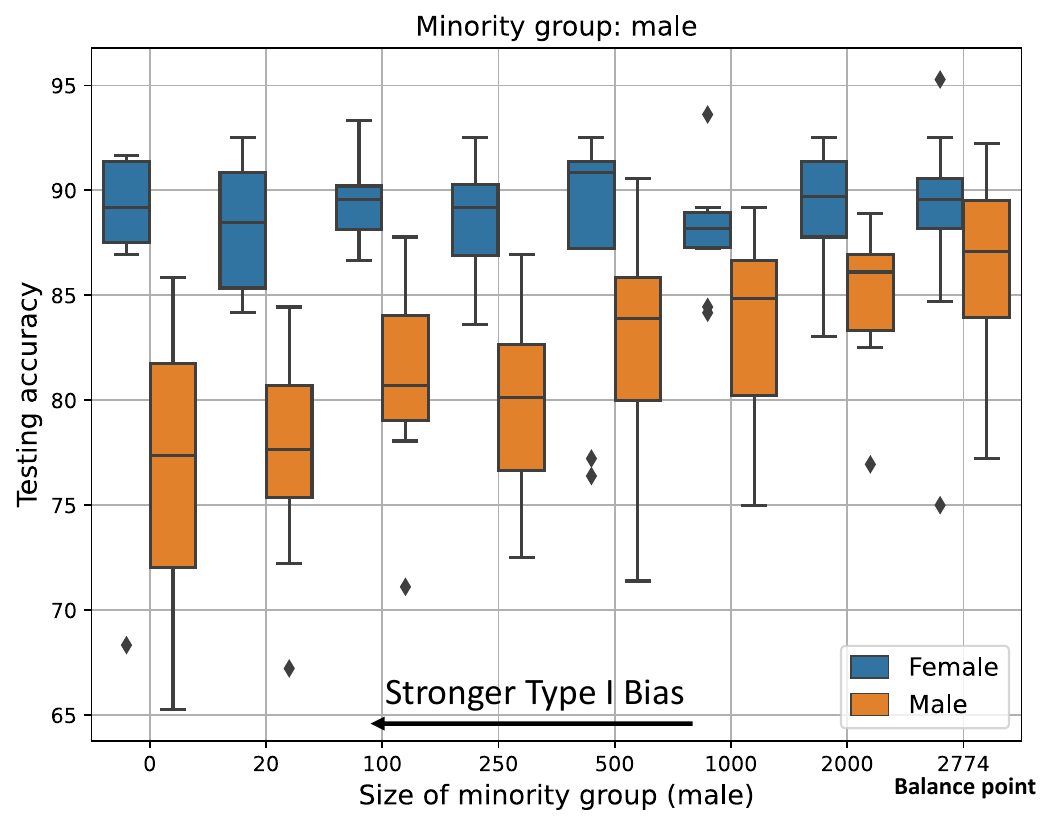}\label{fig:CelebA_I_male}}\quad
      
      \subfloat[Evaluation with various bias assessment metrics.]{\includegraphics[width=0.9\linewidth]{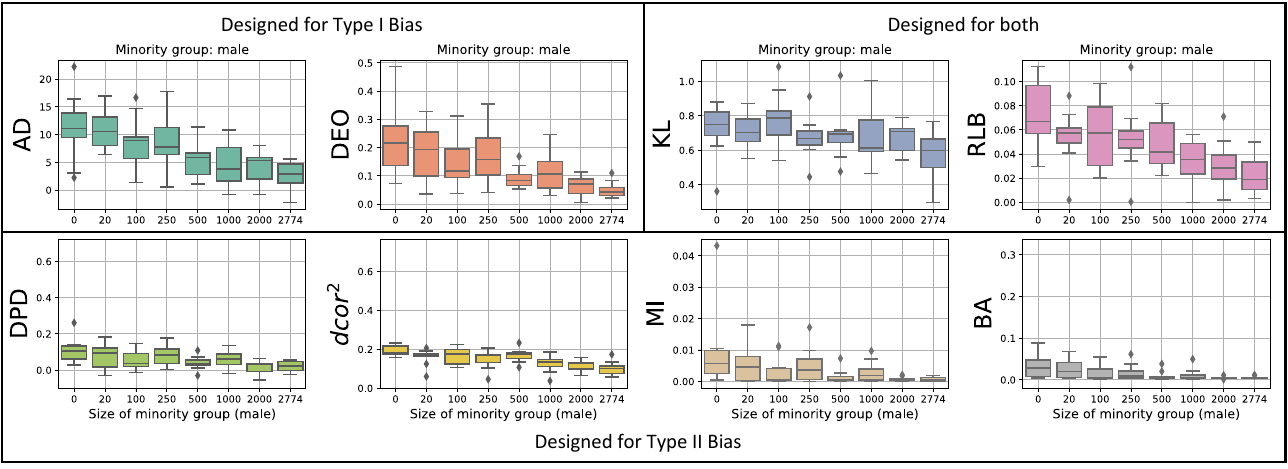}\label{fig:CelebA_metrics_I_male}}\quad

\caption{Investigation of Type I Bias on CelebA with males as minority group. As bias strength diminishes (the size of minority group enlarges), the accuracy of minority group enhances, leading to a reduction in the accuracy disparity between females and males, and the bias assessed by metrics tailored to evaluate Type I Bias is also mitigated.}

\label{fig:CelebA_I}
\end{figure*}

\subsection{Evaluation of Various Metrics}
In this section, we employ CelebA dataset~\cite{CelebA} to investigate several representative bias assessment metrics in assessing Type I Bias and Type II Bias. 
CelebA dataset is an image dataset of human faces where facial attributes (\eg blond hair) are prediction target $Y$ and sex is attribute $A$.

\noindent
\textbf{Setup.}
To construct training and testing sets, we follow the setup of Adult explained above.
In the case of Type I Bias, we construct several training sets with varying bias strength by modifying the size of the minority group in training set.
For testing, we construct a testing set that is balanced across both target and attribute.
In the case of Type II Bias, we construct training sets where facial attributes are associated with a particular sex.
Specifically, we construct an extreme bias version of training set consisting of 89754 images with $H(Y|A)=0$, denoted \textit{TrainEx}, where the bias-conflicting samples (samples exhibiting the opposite bias in training set) are removed from the original training set.
Furthermore, we control bias strength by adjusting the proportion of bias-conflicting samples while maintaining the number of biased samples (samples exhibiting the same bias observed in training set).
For testing, we construct two testing sets: (1) \emph{Unbiased} consisting of 720 images which contain the even number of samples across all combinations of target and attribute, and (2) \emph{Bias-conflicting} consisting of 360 images where all biased samples are excluded from \emph{Unbiased} testing set (only bias-conflicting samples remain).
In both studies, we consider \emph{blond hair} as the prediction target.
For the training model, we utilize ResNet18~\cite{ResNet} followed by the binary cross-entropy loss as the baseline classifier without any debiasing techniques. 
For bias assessment, we employ a comprehensive list of representative metrics including accuracy disparity (AP)~\cite{Accuracy_parity}, difference in equality of opportunity (DEO)~\cite{SensitiveNets}, KL-divergence between score distributions (KL)~\cite{divergence_between_score_distributions}, representation-level bias (RLB)~\cite{RLB}, demographic parity distance (DPD)~\cite{DP_FFVAE}, distance correlation (${dcor}^2)$~\cite{distance_correlation}, mutual information (MI)~\cite{CAT}, and bias amplification (BA)~\cite{BA,Directional_BA}.

\begin{figure*}[htbp]
    \centering %
    
      \subfloat[Testing accuracy.]{\includegraphics[width=0.5\linewidth]{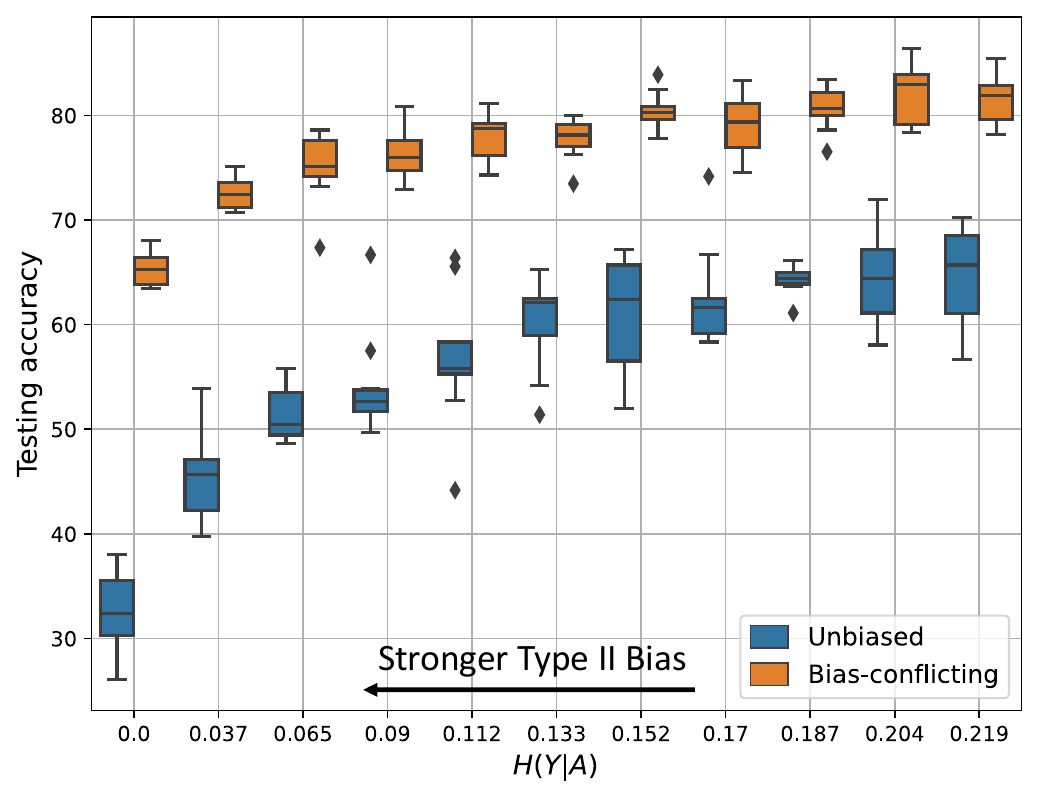}\label{fig:CelebA_II_unbiased}}\quad
      
      \subfloat[Evaluation with various bias assessment metrics.]{\includegraphics[width=0.9\linewidth]{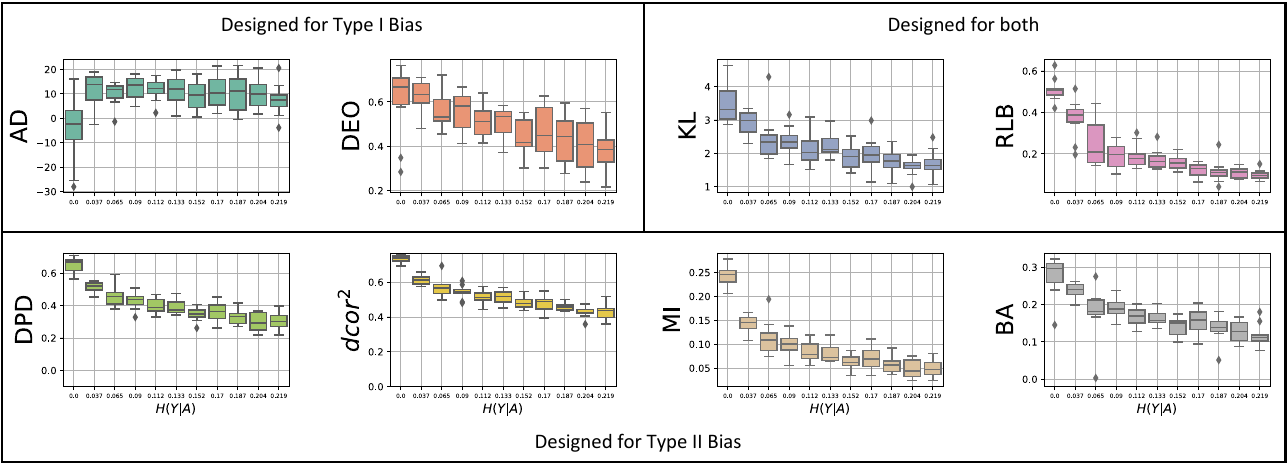}\label{fig:CelebA_metrics_II_unbiased}}\quad

\caption{Investigation of Type II Bias on CelebA. The evaluation of bias assessment metrics is conducted on \emph{unbiased} testing set. As bias strength diminishes ($H(Y|A)$ increases, rendering the attribute less predictive of the target), the accuracies of both \emph{unbiased} and \emph{bias-conflicting} enhance, and the bias assessed by metrics tailored to evaluate Type II Bias is also mitigated.}

\label{fig:CelebA_II}
\end{figure*}

\noindent
\textbf{Analysis}
In the case of Type I Bias, as shown in~\cref{fig:CelebA_I_male}, there exists a noticeable performance disparity across sex. 
As the size of minority group increases (bias strength diminishes), the performance of the minority group improves and the performance gap between the minority and majority groups is mitigated.
Notably, the performance gap is nonzero even at the balance point, with females achieving higher accuracy than males. 
We hypothesize that this is because blond hair is more visually prominent in females with long hair.
Consequently, even if the dataset is balanced across sex, males may be still relatively underrepresented, \ie male images are still insufficient for the model to learn a robust representation of males.
In the case of Type II Bias, as shown in~\cref{fig:CelebA_II_unbiased}, the testing accuracy of both \emph{Unbiased} and \emph{Bias-conflicting} testing set rises as $H(Y|A)$ increases (bias strength diminishes).

For the evaluation of various bias assessment metrics, in~\cref{fig:CelebA_metrics_I_male,fig:CelebA_metrics_II_unbiased}, we observe a noticeable decline in the metrics tailored for a specific type of bias as the corresponding bias strength diminishes.
It is noteworthy that the mean of accuracy disparity (AD) approaches zero in the extreme bias case of Type II Bias where $H(Y|A)=0$ (the leftmost point).
This can be attributed to the fact that, in such extreme bias situations, the target label is bijectively mapped to the attribute label in the training set. Consequently, the trained model may output arbitrary predictions for both sex in the testing set, which leads to an accuracy disparity that is nearly zero.

\section{Path to Follow}
\label{sec:following}

In this section, we present a more comprehensive comparison between Type I Bias and Type II Bias based on our investigation of \pc papers.
Our comparison encompasses multiple aspects including the underlying causes, debiasing methods, evaluation protocol, prevalent datasets, and future directions.
Most notably, for each type of bias, we summarize debiasing methods in~\cref{tab:debiasing_methods}, bias assessment metrics in~\cref{tab:bias_assessment_metrics}, and prevalent datasets in~\cref{tab:datasets_I,tab:datasets_II}.
We hope the comparison can alleviate the cognitive burden from the prevailing confusion between these two types of biases and serve as a roadmap for new researchers to follow.

\begin{table*}[htbp]
\centering
   \caption{The summary of debiasing methods.}
\label{tab:debiasing_methods}

\begin{tabular}{llll}
\toprule
Category     & Pre-processing                                                                                           & In-processing                                                                & Post-processing                                   \\
\midrule
Type I Bias  & Balanced dataset collection~\cite{Timnit_sex_classification_PPB,Fairface}                                & Domain adaptation~\cite{RFW_IMAN,MFR,BAE}                                    & Calibrated equalized odds~\cite{calibrated_Eodds} \\
             & Synthetic dataset generation~\cite{transect, CAT}                                                        & Attribute removal~\cite{DebFace,pass}                                        &                                                   \\
             & Strategic sampling or reweighting~\cite{RL_RBN}                                                                         &                                                                              &                                                   \\
             \midrule
Type II Bias & Universal dataset collection~\cite{extreme_bias} & Mutual information minimization~\cite{learn_not_to_learn_Colored_MNIST,Back_MI,CSAD} & Ensemble domain-independent training~\cite{DI}    \\
             & Synthetic dataset generation~\cite{DP_difference_fpr_GAN_debiasing,fairnessgan_DP_difference_error_rate} & Domain-invariant learning~\cite{Group_DRO, PGI_invariant, EIIL} &                                                  \\
                                                                & Domain randomization~\cite{domain_randomization} & Adversarial training~\cite{LfF_CelebA_Bias_conflicting,BlindEye_IMDB_eb,gradient_projection} & \\
             \bottomrule
\end{tabular}
\end{table*}

\begin{table*}[htbp]
\centering
   \caption{The summary of bias assessment metrics.}
   \label{tab:bias_assessment_metrics}

\begin{tabular}{lll}
\toprule
Category     & Metrics                                                                                                                                                                                             &  \\
\midrule
Type I Bias  & Difference in performance evaluated by various criteria (\eg accuracy disparity (AD)~\cite{multiaccuracy, Accuracy_parity, disparate_mistreatment_on_FPR, conditional_learning}) &  \\
             & Difference in equality of opportunity (DEO)~\cite{SensitiveNets, HSIC, fairnessgan_DP_difference_error_rate, Fairalm_DP_difference_false_positive_rate, DP_difference_fpr_GAN_debiasing}            &  \\
             & Equal error rate (EER)~\cite{FlowSAN}                                                                                                                                                               &  \\
             \midrule
Type II Bias & Demographic parity distance (DPD)~\cite{DP_FFVAE,multiaccuracy,fairnessgan_DP_difference_error_rate}                                                                                                &  \\
             & Distance correlation (${dcor}^2$)~\cite{distance_correlation, BR_Net_dataset_vs_task}                                                                                                                &  \\
             & Mutual information (MI)~\cite{CAT}                                                                                                                                                                  &  \\
             & Bias amplification (BA)~\cite{DI, DP_difference_fpr_GAN_debiasing}, Directional BA~\cite{Directional_BA, DP_difference_fpr_GAN_debiasing}, Multi-attribute BA~\cite{MDBA}
                                           &  \\
             & Disparity impact~\cite{fairness_constraints,AIF360}                                                                                                                                                 &  \\
             & Representation bias~\cite{Resound, Repair}                                                                                                                                                          &  \\
             & Logit-level loss~\cite{CAI_rz,UAI_rz}                                                                                                                                                               &  \\
             \midrule
Both         & KL-divergence between score distributions (KL)~\cite{divergence_between_score_distributions,DP_difference_fpr_GAN_debiasing}                                                                        &  \\
             & Representation-level bias (RLB)~\cite{RLB}                                                                                                                                                          & \\
             \bottomrule
\end{tabular}
\end{table*}

\subsection{Type I Bias} 

\subsubsection{Underlying causes}
Data imbalance across different demographic groups in the training set is commonly accepted as the possible cause for Type I Bias~\cite{cherepanova2023deep,roosli2022peeking}.
Specifically, real-world data often exhibits the long-tail distribution where some demographic groups yield fewer samples than other groups~\cite{long_tail}. 
Consequently, given the data-driven nature of neural networks, models may be effectively trained in groups with sufficient samples but undertrained in groups only with limited samples, hence resulting in performance disparity across different groups and lower performance for minority groups.
On the other hand, recent work suggests that Type I Bias can manifest even when the training set is balanced across demographic groups~\cite{RL_RBN}.
This challenges the conventional understanding of the causes of Type I Bias but promotes the discussion of other possible causes. For instance, Type I Bias may be induced by the underrepresentation of specific demographic groups~\cite{spurious_correlation_Underrepresentation} or the intrinsic challenges associated with recognizing and classifying specific demographic groups~\cite{FR_inherent_bias}.

\subsubsection{Debiasing methods}
Addressing Type I Bias essentially involves optimizing the model to enhance its performance for minority groups while maintaining its performance for majority groups.
The strategies can be broadly classified into three main categories based on the stage when the debiasing intervention is applied relative to the model training phase: pre-processing, in-processing, and post-processing.
First, pre-processing methods intervene before the training phase. 
They are primarily designed based on the cause of Type I Bias (the imbalanced distribution across demographic groups in the training set).
For instance, the straightforward approach is to construct a balanced real dataset for training~\cite{Fairface} or supplement minority groups with sufficient synthetic training samples~\cite{CAT}.
Another approach in this category involves strategically resampling to increase the occurrence of samples from minority groups or reweighting to assign higher importance to samples from underrepresented groups~\cite{RL_RBN}.
Second, in-processing methods are integrated during the model training phase.
Most notably, domain adaptation techniques~\cite{BAE,MFR} adapt well-learned representations from the majority group to the minority group; and, attribute removal methods leverage adversarial learning~\cite{DebFace,pass} to remove demographic information from learned representation.
Lastly, post-processing methods apply debiasing techniques after the training process.
One common technique is to calibrate the model predictions, ensuring that they adhere to specific fairness criteria (\eg equalized odds)~\cite{calibrated_Eodds}.

\subsubsection{Evaluation protocol}
The effectiveness of methods addressing Type I Bias is evaluated by performance disparity between majority and minority groups. 
In the case of binary attributes, the disparity is directly gauged by performance difference between majority and minority groups~\cite{Timnit_sex_classification_PPB,DP_difference_fpr_GAN_debiasing,FPR_Penalty_Loss}.
In the case of non-binary attributes, the disparity is gauged by the standard deviation of performance across all demographic groups (STD)~\cite{DB_VAE_algorithmic_bias, DebFace, GAC, Asymmetric_Rejection_Loss}.
To assess performance, there are a variety of metrics such as error rate~\cite{Timnit_sex_classification_PPB,fairnessgan_DP_difference_error_rate}, loss~\cite{representation_disparity}, accuracy~\cite{multiaccuracy}, average precision (AP)~\cite{DP_difference_fpr_GAN_debiasing}, positive predictive value (PPV), true positive rate (TPR)~\cite{pass,BR_Net_dataset_vs_task}, false positive rate (FPR)~\cite{FPR_Penalty_Loss}, average false rate (AFR), mean AFR (M AFR)~\cite{inclusivefacenet}, confusion matrix~\cite{DebFace}, F1 score~\cite{BR_Net_dataset_vs_task}, receiver operating characteristic curve (ROC)~\cite{RL_RBN, fairnessgan_DP_difference_error_rate, SAN, Asymmetric_Rejection_Loss,debias_balanced_AUCROC}, area under the ROC (AUC)~\cite{FlowSAN, DebFace, BR_Net_dataset_vs_task}.
Furthermore, besides these metrics to assess performance disparity, the performance improvement in minority groups compared to the baseline is provided for an intuition of debiasing effectiveness, along with overall performance to illustrate that it is not compromised.

\subsubsection{Datasets}
Datasets used to investigate Type I Bias mainly exhibit long-tail distributions.
Most notably, several benchmark biometric datasets including LFW~\cite{LFW}, IJB-A~\cite{IJBA}, IJB-C~\cite{IJBC}, and RFW \cite{RFW_IMAN}, are frequently utilized.
A comprehensive list of datasets is presented in~\cref{tab:datasets_I}.

\begin{table*}[htbp]
\centering
   \caption{The well-known datasets used to study Type I Bias.}
   \label{tab:datasets_I}
\begin{tabular}{lccccccccc}
\toprule
\multirow{2}{*}{Name} & \multirow{2}{*}{Subjects} & \multirow{2}{*}{Images} & \multicolumn{2}{c}{Sex (\%)} & \multicolumn{5}{c}{Race (\%)}                                                                         \\  
\cmidrule(lr){4-5}  \cmidrule(lr){6-10} 
                      &                           &                         & Female           & Male          & European             & Asian            & Indian            & African            & Hispanic or Latino            \\ 
                      \midrule
CelebA~\cite{CelebA}                & 10K                       & 202.5K                  & 58.3             & 41.7          & -                    & -                & -                 & -                  & -                   \\ 
MUCT~\cite{MUCT}                  & 0.2K                      & 3.7K                    & 50.9             & 49.1          & -                    & -                & -                 & -                  & -                   \\ 
RaFD~\cite{RaFD}                  & 67                        & 1.6K                    & 37.3             & 62.7          & -                    & -                & -                 & -                  & -                   \\ 
PPB~\cite{Timnit_sex_classification_PPB}                   & 1.2K                      & 1.2K                    & 44.6             & 55.4          & 48.0                 & -                & -                 & 52.0               & -                   \\ 
MORPH~\cite{MORPH}                 & 13.6K                     & 55.1K                   & 15.3             & 84.7          & 19.2                 & 0.28             & -                 & 77.2               & 3.2                 \\ 
LFW~\cite{LFW}                   & 5.7K                      & 13K                     & 22.3             & 77.6          & 69.9                 & 13.2             & 2.9               & 14.0               & -                   \\ 
CASIA-Webface~\cite{webface}         & 10K                       & 0.5M                    & 58.9             & 41.1          & 84.5                 & 2.6              & 1.6               & 11.3               & -                   \\ 
VGGFace2~\cite{vggface2}              & 8.6K                      & 3.1M                    & 59.3             & 40.7          & 74.2                 & 6.0              & 4.0               & 15.8               & -                   \\
MS-Celeb-1M~\cite{MS_celeb_1m}           & 90K                       & 5.0M                    & -                & -             & 76.3                 & 6.6              & 2.6               & 14.5               & -                   \\ 
IJB-A~\cite{IJBC}                 & 0.5K                      & 5.7K                    & -                & -             & 66.0                 & 9.8              & 7.2               & 17.0               & -                   \\ 
IMDB-WIKI~\cite{IMDB_WIKI}             & 20K                          & 500K                    & 41.1                 & 57.1              & 79.5                 & 2.6              & 2.3               & 11.5               & 4.1                 \\ 
UTK~\cite{UTKFace}                   & -                          & 20K                     & \multicolumn{2}{c}{Balanced}               & 45.3                 & 14.7             & 18.4              & 21.6               & -                   \\ 
RFW~\cite{RFW_IMAN}                   & 12K                       & 40K                     & 27.7                & 72.3             & \multicolumn{4}{c}{Balanced}                & -                   \\ 
FairFace~\cite{Fairface}              & -                      & 108K                    & \multicolumn{2}{c}{Balanced}                       & \multicolumn{5}{c}{Balanced} \\ 
\bottomrule
\end{tabular}
\end{table*}

\subsubsection{Future directions}
One promising future direction is to delve into the root cause of Type I Bias since the formerly widely accepted cause (data imbalance) has been challenged by the experiment that Type I Bias exists even for a balanced dataset~\cite{RL_RBN}.
Furthermore, exploring more effective debiasing methods to achieve even performance across cohorts is always of significant importance, hence it is a valuable direction.

\subsection{Type II Bias}

\subsubsection{Underlying causes}
The association between prediction targets and attributes in the training set is widely considered the possible cause of Type II Bias~\cite{LfF_CelebA_Bias_conflicting,CSAD,ECS}. 
Different from Type I Bias, which originates from an uneven distribution of samples across attributes, Type II Bias arises from an uneven distribution of \emph{specific target groups} across attributes. 
Specifically, the collected data may encompass a greater number of samples annotated with specific pairs of target labels and attribute labels (\eg $(y^1,a^1)$ and $(y^2,a^2)$) than other combinations. 
Models trained on this dataset may leverage these attributes as shortcut features to simplify the training process rather than acquiring more comprehensive features. 
Consequently, when applying the trained models in real-world scenarios where the association does not generally exist, they may still rely on these attributes for decision-making and yield predictions that depend on these attributes, thereby resulting in a higher frequency of particular prediction outcomes for particular groups and further unfair treatment for these groups.

\subsubsection{Debiasing methods}
Addressing Type II Bias essentially involves acquiring representations that are independent of the attribute while remaining informative for a wide range of downstream tasks~\cite{FNF}.
Similar to Type I Bias, the strategies can be classified into three categories: pre-processing, in-processing, and post-processing.
First, pre-processing approaches can be further sub-categorized into dataset construction and data preprocessing.
Dataset construction mainly encompasses collecting large-scale universal datasets to lessen the likelihood of spurious correlation between the target and the attribute~\cite{extreme_bias, sabaf}, and generating counterfactual synthetic samples to augment the original biased training set, thereby reducing its inherent bias strength~\cite{CGN,BiaSwap,Camel,DP_difference_fpr_GAN_debiasing}.  
Data preprocessing mainly encompasses fairness through unawareness~\cite{fairness_through_awareness}, which directly eliminates attributes from the input data, and domain randomization~\cite{domain_randomization} to utilize domain knowledge to assign a random value to the attribute label for each sample, thereby rendering it irrelevant to the target prediction. 
Second, in-processing approaches can be further divided into two subgroups: methods that either explicitly or implicitly minimize the mutual information (MI) between the learned latent features and the specific attribute.
Specifically, several methods directly minimize mutual information between the latent representation for the target classification and the protected attributes to learn a representation that is predictive of the target but independent of the attributes~\cite{learn_not_to_learn_Colored_MNIST,Back_MI,CSAD}. 
Another group of methods applies adversarial learning with surrogate losses~\cite{LfF_CelebA_Bias_conflicting,BlindEye_IMDB_eb,gradient_projection} to implicitly reduce the mutual information or utilize domain-invariant learning~\cite{ganin2016domain, zhao2019learning, albuquerque2019generalizing, Group_DRO, PGI_invariant, EIIL} to minimize classification performance gap across groups by mapping data to a space where distributions are indistinguishable while maintaining task-relevant information. 
Lastly, for the post-processing method, domain-independent learning~\cite{DI} learns an ensemble classifier comprising separate classifiers for each demographic group by sharing representations, thereby ensuring that the prediction from the unified model is not biased towards any domain.

\subsubsection{Evaluation protocol}
The effectiveness of methods addressing Type II Bias is evaluated by prediction disparity across different groups.
In the prevalent evaluation protocol, models are trained on a dataset where the target is associated with the attribute and tested on a held-out dataset where such association is absent~\cite{learn_not_to_learn_Colored_MNIST, Back_MI, CSAD}.
Subsequently, the testing accuracy is reported to evaluate the model capability to reduce the effect of association in the training set (the effectiveness to mitigate Type II Bias)~\cite{DI}.
Several studies also present the accuracy of worst-case groups, where the samples yield the opposite of bias present in training set~\cite{Group_DRO, JTT, confused_dataset_bias_DFA}.
Furthermore, we summarize other commonly-used bias assessment metrics in~\cref{tab:bias_assessment_metrics}.
A noteworthy distinction in these bias assessment metrics for Type II Bias compared with Type I Bias is the absence of necessity for ground truth labels. This distinction is attributed to the fact that Type II Bias is defined as the dependence between model prediction and attribute, eliminating the need for ground truth, while evaluating Type I Bias necessitates ground truth to assess model performance.

\subsubsection{Datasets}
Most notably, several census datasets, including Adult income dataset~\cite{adult_dataset_and_german_dataset}, German credit dataset~\cite{adult_dataset_and_german_dataset}, and COMPAS recidivism dataset~\cite{COMPAS}, are employed as benchmark datasets to investigate the impact of sensitive/protected attributes in real-world decision-making processes. 
Additionally, computer vision and natural language processing communities also develop various datasets to investigate Type II Bias, \eg Colored MNIST~\cite{learn_not_to_learn_Colored_MNIST}, CelebA~\cite{CelebA,LfF_CelebA_Bias_conflicting}, Waterbirds~\cite{Group_DRO}, and CivilComments-WILDS~\cite{CivilComments,Wilds}.
A comprehensive list of datasets is summarized in~\cref{tab:datasets_II}.

\begin{table*}[htbp]
\centering
   \caption{The well-known datasets used to study Type II Bias.}
   \label{tab:datasets_II}

\begin{tabular}{llll}
\toprule
Name                                                  & Modality & Attribute              & Target                                \\
\midrule
Adult~\cite{adult_dataset_and_german_dataset}         & Tabular   & Sex                    & Income                                \\
German~\cite{adult_dataset_and_german_dataset}        & Tabular   & Sex, age               & Credit                                \\
COMPAS~\cite{COMPAS}                                  & Tabular   & Race                   & Recidivism                            \\
Colored MNIST~\cite{learn_not_to_learn_Colored_MNIST} & Image    & Color                  & Digit                                \\
CelebA~\cite{CelebA}                                  & Image    & Sex                    & Facial attributes                     \\
IMDB~\cite{IMDB}                                      & Image    & Sex, age               & Age, sex                              \\
Waterbirds~\cite{Group_DRO}                           & Image    & Background             & Waterbirds or landbirds               \\
CivilComment-WILDS~\cite{Wilds}                       & Text     & Demographic identities & Toxic or non-toxic \\
\bottomrule
\end{tabular}

\end{table*}

\subsubsection{Future directions}
One promising research direction is to explore the strong bias region~\cite{extreme_bias} of Type II Bias, where the target and the attribute are strongly associated in the training set, a scenario that is overlooked by many existing work~\cite{BlindEye_IMDB_eb, learn_not_to_learn_Colored_MNIST}.
Also, it is important to further explore more challenging scenarios where attribute labels are absent~\cite{HEX_texture_bias1, ReBias_texture_bias2,rubi} or unknown biases emerge~\cite{discover_unknown_bias,CNC,EIIL}.

\subsection{Summary}
In this section, we highlight the distinctions between Type I Bias and Type II Bias across multiple aspects and provide further explanations on the comparison in~\cref{tab:teaser}.

\begin{itemize}

    \item Manifestation. A model exhibiting Type I Bias yields uneven performance across different groups and lower performance in minority groups, whereas a model exhibiting Type II Bias depends on attributes for decision-making and produces specific predictions that are highly associated with specific attributes. 
    
    \item Disparity. Type I Bias refers to the disparity in prediction performance across attributes, whereas Type II Bias refers to the disparity in prediction outcomes across attributes.
    
    \item Causes. Type I Bias stems from insufficient training of underrepresented groups, whereas Type II Bias arises from the association between targets and attributes.
    
    \item Dataset inducing bias. An imbalanced distribution of samples across attributes induces Type I Bias, whereas an imbalanced distribution of \emph{specific target groups} across attributes induces Type II Bias.
    
\end{itemize}

\section{Suggestions}
\label{sec:suggestion}
In this section, we propose several suggestions to elucidate how researchers engaged in bias-related work can avoid the existing confusion in~\cref{sec:confusion}.
First, we suggest that researchers explicitly and precisely specify the type of bias they address, and avoid vague terminology. 
In this sense, utilizing terminology which is unequivocally defined, \eg Type I Bias and Type II Bias, will provide clear and undisputed information.
Second, we recommend that researchers derive motivation for their own work from the work that addresses the identical type of bias.
By doing this, the existing confusion can be gradually diminished.
Third, we advise researchers to abstain from introducing new terminology for previously discussed biases, and clarify the difference between previous definitions and the newly proposed definition if the new definition is necessary.
Hereby, the reuse of established terms will help foster a clear and unified community.

\section{Conclusion}
\label{sec:conclusion}
Through an investigation of \pc papers, we uncover the substantial confusion, surrounding two prevalent types of biases within the machine learning community, which amplifies the learning burden for new researchers.
Subsequently, we delve into the possible causes of the confusion.
Most notably, we observe that researchers from diverse backgrounds hold different preconceptions about bias, leading to a lack of unified terminology for the same type of bias over an extended period.
To alleviate the existing confusion and restore clarity in the literature, we present mathematical definitions for these two prevalent types of biases. 
Furthermore, we unify a comprehensive list of papers under these definitions and distinguish these two types of biases from multiple perspectives.
Through this endeavor, we seek to facilitate the discussion on bias-related issues among researchers with diverse backgrounds.

\bibliographystyle{IEEEtran}
\bibliography{reference}

\section{Biography Section}
\vspace{-12mm}

\begin{IEEEbiography}[{\includegraphics[width=1in,height=1.25in,clip,keepaspectratio]{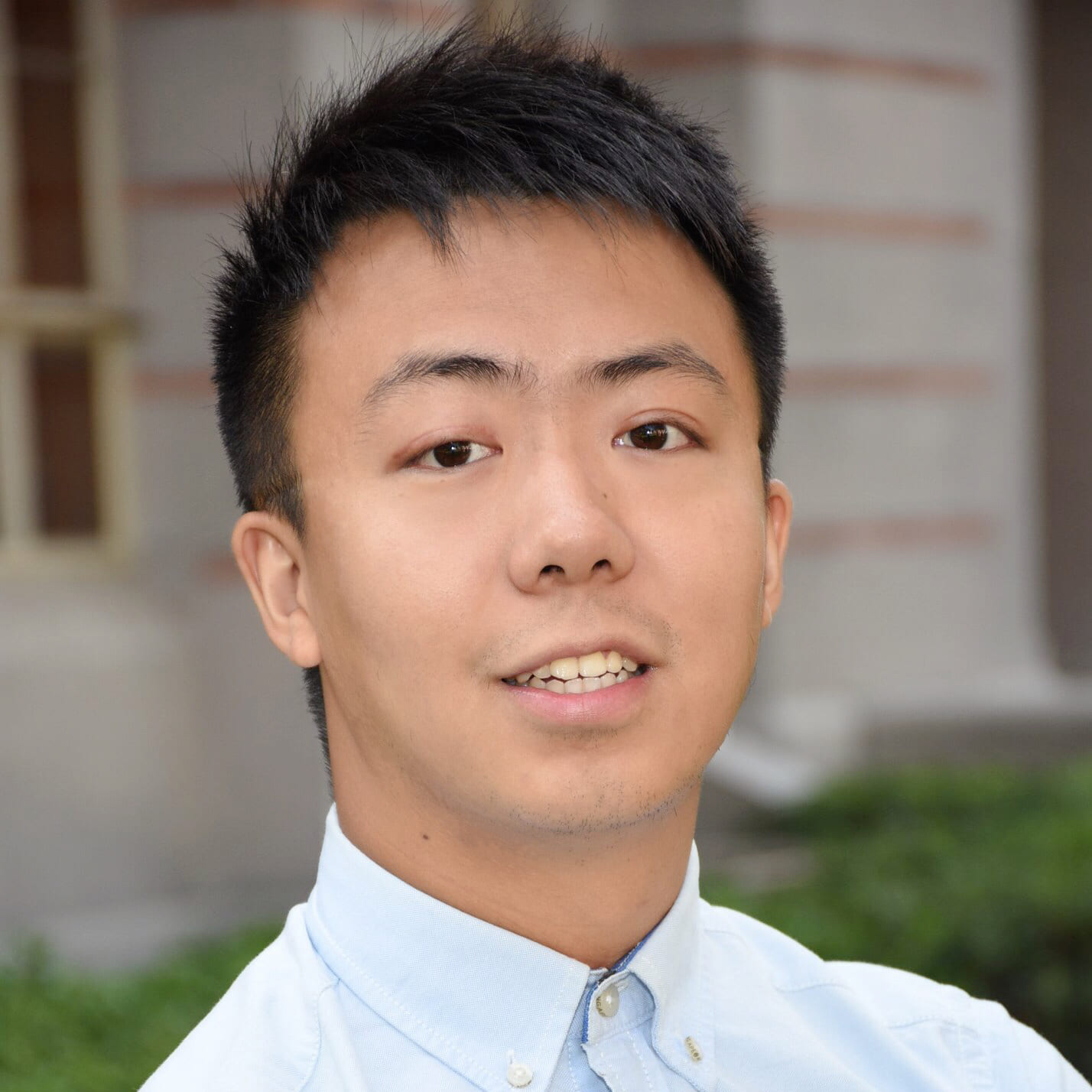}}]{Jiazhi Li}
received his Bachelor's degree in Electrical Engineering from Beijing Institution of Technology in Beijing, China in 2018 and his Master's degree in Electrical Engineering from University of Southern California, USA in 2020. Currently, he is a Ph.D. student at the Department of Electrical and Computer Engineering, and a Graduate Research Assistant at Information Sciences Institute, both being units of USC Viterbi School of Engineering, under the supervision of Prof. Jieyu Zhao and Prof. Wael AbdAlmageed. His research interests include machine learning fairness and generative models.
\end{IEEEbiography}

\vspace{-12mm}

\begin{IEEEbiography}[{\includegraphics[width=1in,height=1.25in,clip,keepaspectratio]{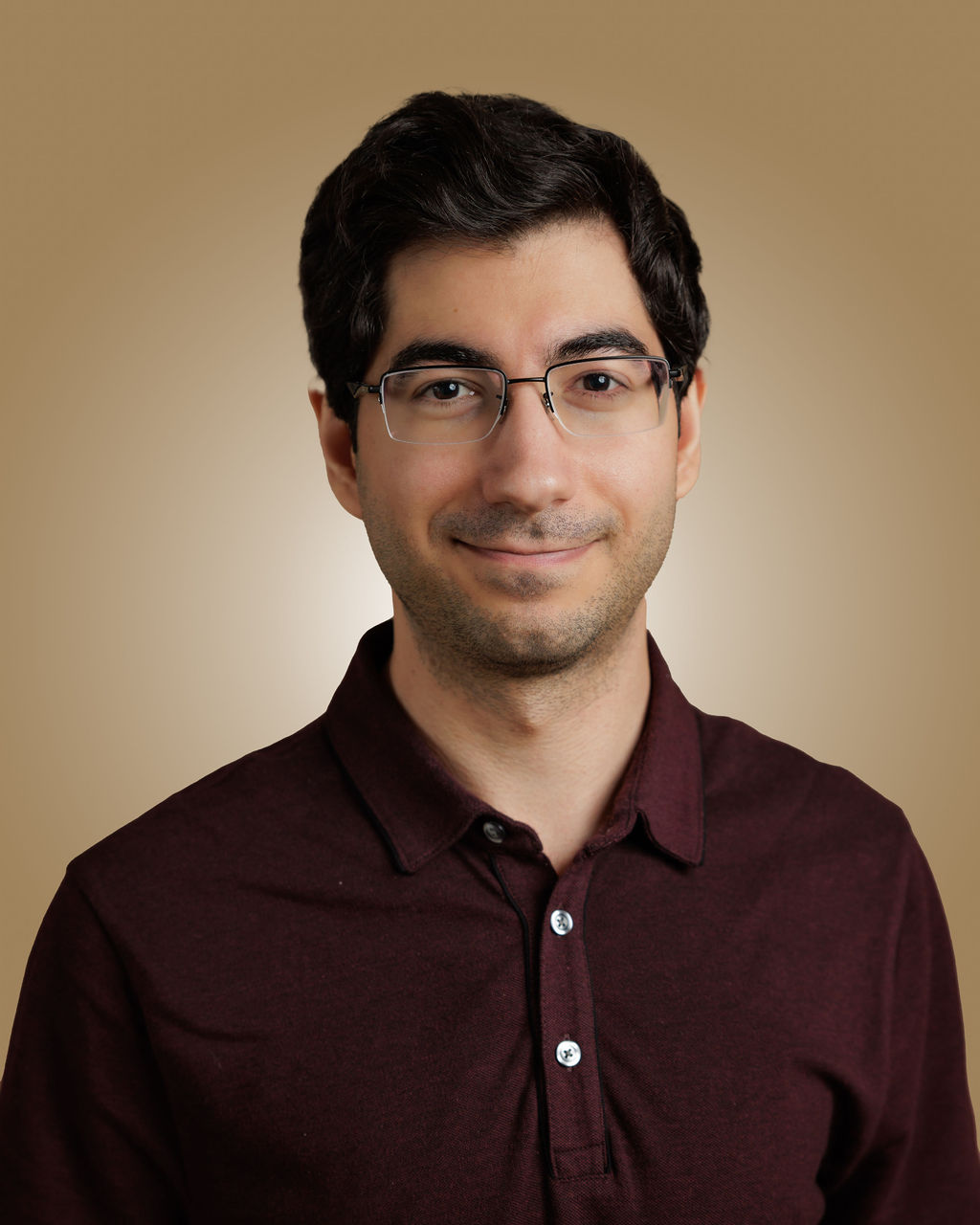}}]{Mahyar Khayatkhoei}
is a Computer Scientist at USC Information Sciences Institute. He received his B.Sc. in electrical engineering from the University of Tehran, and his M.Sc. and Ph.D. in computer science from Rutgers University. His research focuses on identifying and measuring the biases and limitations of deep neural networks in general, and the theory and application of deep generative models in particular.
\end{IEEEbiography}

\vspace{-14mm}

\begin{IEEEbiography}[{\includegraphics[width=1in,height=1.25in,clip,keepaspectratio]{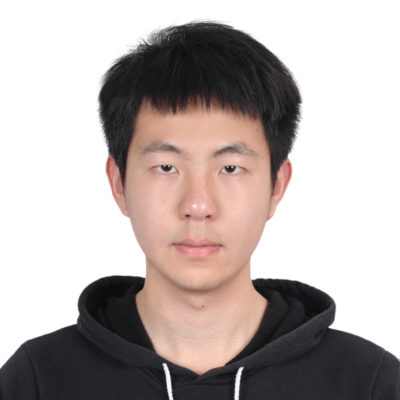}}]{Jiageng Zhu}
is a Ph.D. student at USC Ming Hsieh Department of Electrical and Computer Engineering and a Graduate Research Assistant at USC Information Sciences Institute. His current research interests focus on Causal Representation Learning, Disentanglement and Invariant Representation Learning, and Dynamics Prediction.
\end{IEEEbiography}

\vspace{-18mm}

\begin{IEEEbiography}[{\includegraphics[width=1in,height=1.25in,clip,keepaspectratio]{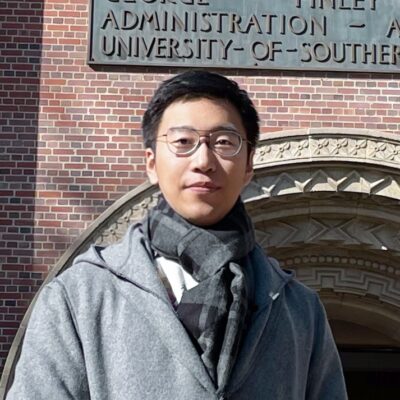}}]{Hanchen Xie}
is a Ph.D. candidate at USC Thomas Lord Department of Computer Science and a Graduate Research Assistant at USC Information Sciences Institute; both are units of USC Viterbi School of Engineering. His research interests include representation learning under less labeled data scenarios (\eg Semi-Supervised Learning, Few-Shot Learning, and Zero-Shot Learning), generative networks, and Dynamics Prediction.
\end{IEEEbiography}

\vspace{-15mm}

\begin{IEEEbiography}[{\includegraphics[width=1in,height=1.25in,clip,keepaspectratio]{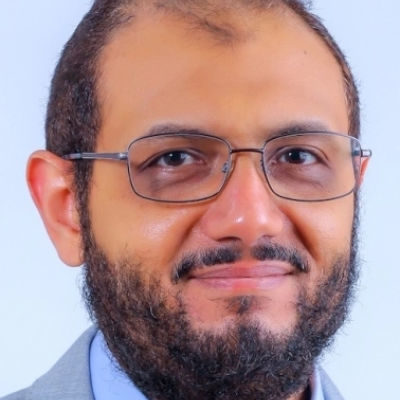}}]{Mohamed E. Hussein}
is a Computer Scientist and Research Lead at USC ISI, and an Associate Professor (on leave) at Alexandria University, Egypt. He obtained his Ph.D. degree in Computer Science from the University of Maryland at College Park in 2009, specializing in computer vision and GPU computing.
His current research interest is in mitigating the vulnerabilities of AI models to spoofing attacks, adversarial attacks, and domain shifts.
\end{IEEEbiography}

\vspace{-15mm}

\begin{IEEEbiography}[{\includegraphics[width=1in,height=1.25in,clip,keepaspectratio]{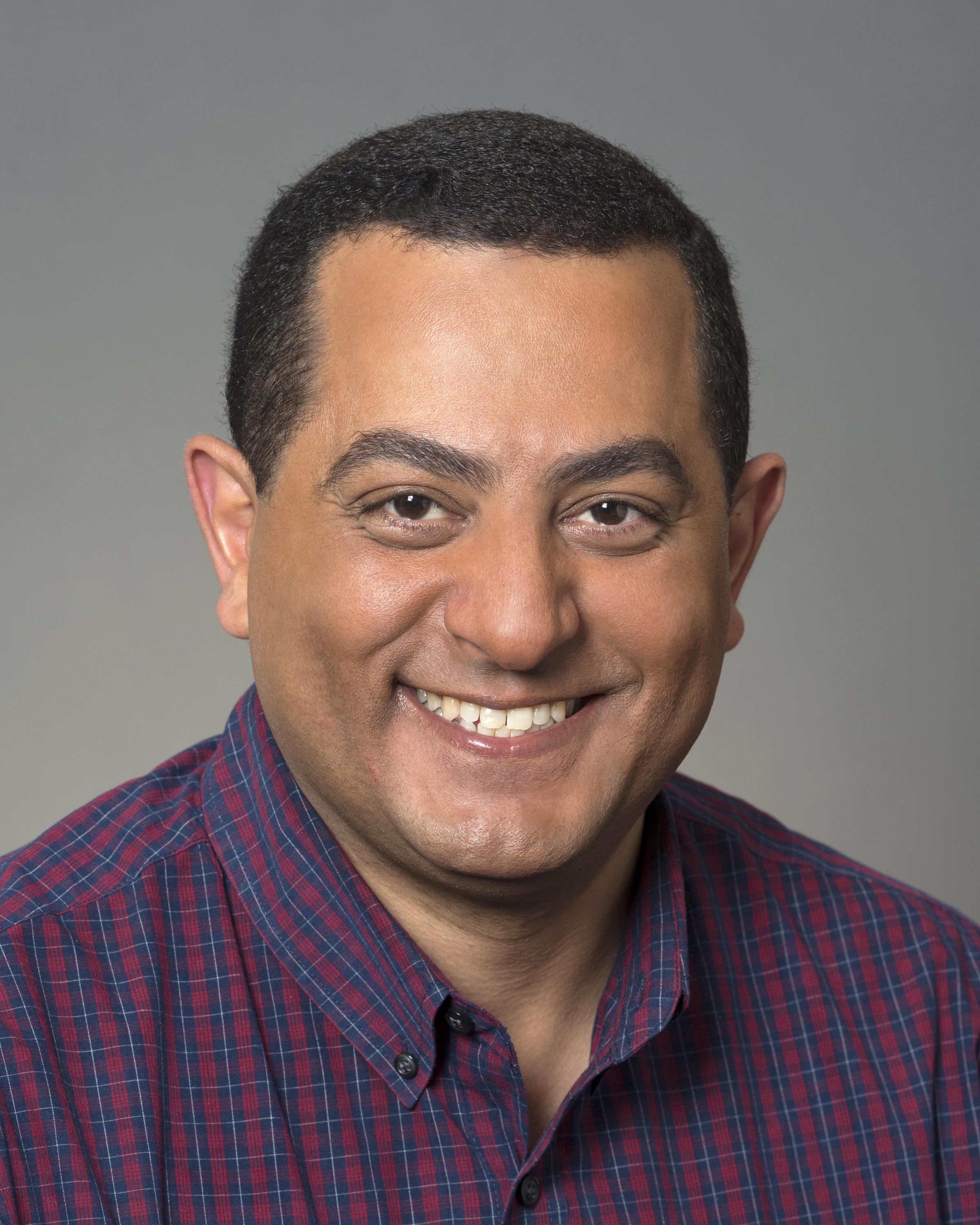}}]{Wael AbdAlmageed}
is a Tenured Full Professor at the Holcombe Department of Electrical and Computer Engineering at Clemson University. From 2013 to 2023, he was a Research Associate Professor at Department of Electrical and Computer Engineering, and a Research Director and Distinguished Principal Scientist at Information Sciences Institute. He is the Founding Director of the USC’s Visual Intelligence and Multimedia Analytics Laboratory (VIMAL). He received his B.S. in electrical engineering in 1994 and his M.S. in computer engineering in 1997 from Mansoura University in Egypt. 
He obtained his Ph.D. with Distinction from the University of New Mexico in 2003 where he was also awarded the Outstanding Graduate Student award. His research interests include representation learning, debiasing and fair representations, multimedia forensics and visual misinformation identification (such as deepfake and image manipulation detection), and face recognition and biometric anti-spoofing. He leads several multi-institution research efforts, including DARPA’s MediFor, GARD and LwLL and IARPA’s Janus, Odin and BRIAR. 

\end{IEEEbiography}

\clearpage
\section*{Appendix}
\setcounter{section}{0} 
\renewcommand\thesection{\arabic{section}}
\label{sec:related_work}

\section{Full categorization}
In this section, we provide a comprehensive list of all \pc papers that investigate Type I Bias and Type II Bias. 
Besides, for more fine-grained categorization, we classify papers addressing predominant issues into various subgroups.

\subsection{Type I Bias}

\subsubsection{Biometrics}
\cite{RL_RBN}; 
\cite{RFW_IMAN}; 
\cite{GAC}; 
\cite{MvCoM}; 
\cite{FPR_Penalty_Loss};
\cite{FairCal};
\cite{Asymmetric_Rejection_Loss};
\cite{inclusivefacenet};
\cite{debias_data_augmentation};
\cite{debias_balanced_AUCROC};
\cite{mcduff2019characterizing};
\cite{long2015learning};
\cite{dam};
\cite{dhar2020adversarial};
\cite{terhorst2020post};
\cite{Terhrst2020ComparisonLevelMO};
\cite{robinson2020face};
\cite{sensitive_loss};
\cite{CAT};
\cite{pahl2022female};
\cite{conti2024assessing}.

\noindent
\text{Investigation of the role of demographic information:}
\cite{FR_inherent_bias};
\cite{pass};
\cite{DebFace};
\cite{SensitiveNets};
\cite{personalization};
\cite{class_imbalance_long_tail}; 
\cite{albiero2020face};
\cite{cavazos2020accuracy};
\cite{han2017heterogeneous};
\cite{nagpal2019deep};
\cite{FVRT3}.

\subsubsection{Classification of protected attribute}
\cite{Timnit_sex_classification_PPB}; 
\cite{transect};
\cite{Fairface};
\cite{MTCNN};
\cite{weerts2023algorithmic};
\cite{de2019does};
\cite{cruz2024unprocessing}.

\subsubsection{Other tasks associated with protected attribute}
\cite{representation_disparity}; 
\cite{Fair_robust_learning};
\cite{MSFDA};
\cite{data_feedback};
\cite{FairPCA};
\cite{CLMing};
\cite{fairfil};
\cite{tan2019assessing};
\cite{ULPL};
\cite{adaptive_weights};
\cite{BB_BC};
\cite{rahmattalabi2019exploring};
\cite{NEURIPS2018_cc4af25f}; 
\cite{yurochkin2019training};
\cite{kim2020fair};
\cite{shankar2017no};
\cite{li2014learning};
\cite{xiao2023name};
\cite{bell2023simplicity};
\cite{shrestha2023help};
\cite{mccradden2023s};
\cite{gardner2023cross};
\cite{hutiri2022bias};
\cite{pastaltzidis2022data};
\cite{markl2022language};
\cite{donahue2022human};
\cite{wang2022towards};
\cite{fong2021fairness};
\cite{mclaughlin2022fairness};
\cite{steed2021image};
\cite{dhamala2021bold};
\cite{martin2023bias};
\cite{nanda2021fairness};
\cite{sweeney2020reducing};
\cite{yang2020towards};
\cite{green2019disparate};
\cite{belem2024are}.

\subsubsection{Equalized odds}
\cite{FSCL};
\cite{FURL_PS};
\cite{von_Mises_Fisher};
\cite{Fair_consistency_regularization};
\cite{shift_structure};
\cite{PAC_learning};
\cite{golz2019paradoxes};
\cite{pmlr-v80-kallus18a};
\cite{calibrated_Eodds};
\cite{kim2020fair};
\cite{NIPS2017_250cf8b5};
\cite{loi2022calibration};
\cite{blum2022multi};
\cite{nandy2022achieving};
\cite{wu2022fairness};
\cite{mishler2021fairness};
\cite{wang2021fair};
\cite{taskesen2021statistical};
\cite{coston2020counterfactual};
\cite{canetti2019soft};
\cite{chang2023feature};
\cite{plecko2024mind};
\cite{cherian2024statistical};
\cite{small2024equalised};
\cite{simson2024one}.

\subsubsection{Equal opportunity}
\cite{HSIC};
\cite{CGL};
\cite{POCAR};
\cite{FATDM};
\cite{causal_modeling};
\cite{mismatched_hypothesis_testing};
\cite{discrepancy};
\cite{NEURIPS2018_83cdcec0};
\cite{de2019bias};
\cite{henzinger2023runtime};
\cite{awasthi2021evaluating};
\cite{d2020fairness};
\cite{hu2020fair};
\cite{heidari2019moral};
\cite{kearns2019empirical};
\cite{corbett2017algorithmic};
\cite{tang2024procedural};
\cite{selialia2024mitigating};
\cite{binkyte2024babe}.

\subsubsection{Accuracy parity}
\cite{multiaccuracy};
\cite{Accuracy_parity};
\cite{disparate_mistreatment_on_FPR};
\cite{conditional_learning};
\cite{accuracy_parity_gap};
\cite{FRL};
\cite{mickel2024racial};
\cite{baumann2024fairness};
\cite{akpinar2024impact};
\cite{lunich2024explainable};
\cite{dong2024addressing};
\cite{lee2024large}.

\subsection{Type II Bias}

\subsubsection{Labeled spurious attribute}
\cite{Group_DRO};
\cite{DARE};
\cite{DFR};
\cite{Study_core_features_DFR};
\cite{CVaR_DRO};
\cite{CSAD};
\cite{Back_MI};
\cite{learn_not_to_learn_Colored_MNIST};
\cite{BlindEye_IMDB_eb};
\cite{End};
\cite{BCL};
\cite{selecmix};
\cite{CLGR};
\cite{annotation_bias};
\cite{DI};
\cite{DP_difference_fpr_GAN_debiasing};
\cite{Resound};
\cite{Repair};
\cite{DisC};
\cite{DropClass};
\cite{DVQA};
\cite{ContraCAM};
\cite{rubi};
\cite{idrissi2022simple};
\cite{LearnedMixin};
\cite{Camel};
\cite{CGN};
\cite{adila2024discovering};
\cite{albuquerque2024evaluating};
\cite{jung2024simple};
\cite{zeng2024understanding};
\cite{sreelatha2024denetdm};
\cite{chakraborty2024visual};
\cite{alabdulmohsin2024clip}.

\noindent
\text{Simplicity bias:}
\cite{simplicity_bias_CL};
\cite{SIFER};
\cite{FRR};
\cite{feature_distortion};
\cite{Two_layer_Nets};
\cite{gatmiry2024simplicity};
\cite{tsoy2024simplicity};
\cite{rende2024distributional};
\cite{nguyen2019changing};
\cite{he2024towards};
\cite{chen2024sudden}.

\noindent
\text{Shape and texture bias:}
\cite{Style_Randomization};
\cite{li2021shapetexture};
\cite{functional_entropy};
\cite{mishra2020dqi};
\cite{singh2020don}.

\subsubsection{Unlabeled spurious attribute}
\cite{geirhos2018imagenet};
\cite{HEX_texture_bias1}
\cite{ReBias_texture_bias2};
\cite{clark2019don};
\cite{he2019unlearn};
\cite{rubi};
\cite{clark2020learning};
\cite{utama2020mind}.

\subsubsection{Unknown spurious attribute}
\cite{ECS};
\cite{discover_unknown_bias};
\cite{DebiAN};
\cite{UBNet};
\cite{AFLite_investigation};
\cite{BiaSwap};
\cite{RNF};
\cite{forgettable_examples};
\cite{sanh2021learning};
\cite{lahoti2020fairness};
\cite{pezeshki2021gradient};
\cite{LfF_CelebA_Bias_conflicting};
\cite{confused_dataset_bias_DFA};
\cite{JTT};
\cite{LC};
\cite{EIIL};
\cite{CNC};
\cite{SSA}; 
\cite{PGI_invariant};
\cite{CIM};
\cite{George};
\cite{PGD};
\cite{CDvG};
\cite{LWBC};
\cite{Unlabeled_DFR};
\cite{FairKL};
\cite{basumitigating}.

\subsubsection{Labeled sensitive attribute}
\cite{discrimination_score};
\cite{machine_bias};
\cite{deepmed};
\cite{BAF};
\cite{bounded_exploration};
\cite{medfair};
\cite{RDIA};
\cite{CSN};
\cite{AUTOREGRESSIVE};
\cite{Causal_Mediation_Analysis};
\cite{maximum_entropy};
\cite{NIPS2016_a486cd07};
\cite{sadeghi2019global};
\cite{Renyi};
\cite{sun2019mitigating};
\cite{mei2023bias};
\cite{wolfe2023contrastive};
\cite{cabello2023independence};
\cite{bianchi2023easily};
\cite{hirota2022gender};
\cite{ball2021differential};
\cite{cho2021towards};
\cite{obermeyer2019dissecting};
\cite{jung2025unified};
\cite{jung2025counterfactually};
\cite{teo2024fairqueue};
\cite{kim2024training};
\cite{limisiewicz2024debiasing};
\cite{shen2024finetuning};
\cite{fairerclip}.

\subsubsection{Unknown sensitive attribute}
\cite{SSL_without_demographics};
\cite{buet2022towards};
\cite{lu2024debiasing}.

\subsubsection{Demographic parity} 
\cite{DP_FFVAE};
\cite{fairgan};
\cite{decaf};
\cite{CFQP};
\cite{Fair-Greedy};
\cite{EXP_ELIM};
\cite{OTF};
\cite{Dynamic_programming_fair};
\cite{Disparate_treatment};
\cite{FRAUD-Detect};
\cite{R2B};
\cite{DP_postprocessing};
\cite{minimax_fairness};
\cite{fare};
\cite{fairGBM};
\cite{UDDIA};
\cite{Shifty};
\cite{sufficiency_rule};
\cite{auditing};
\cite{RTO};
\cite{structured_prediction};
\cite{Wasserstein_Barycenters};
\cite{MMD_sinkhorn_divergence};
\cite{FR_Train};
\cite{gordaliza2019obtaining};
\cite{zhao2022inherent};
\cite{locatello2019fairness};
\cite{cunningham2021underestimation};
\cite{fairness_constraints};
\cite{yang2023fairness};
\cite{chen2023personalized};
\cite{rateike2022don};
\cite{ghazimatin2022measuring};
\cite{zhang2021towards};
\cite{jin2024maximal};
\cite{xiong2023fair};
\cite{ohayon2024perceptual};
\cite{defrance2024abcfair};
\cite{NEURIPS2024_142bff4f};
\cite{dehdashtian2024fairerclip};
\cite{liu2024towards};
\cite{kang2024deceptive};
\cite{cachel2024prefair};
\cite{difference_ratio}.

\subsection{Both Type I and Type II Biases}
\cite{DB_VAE_algorithmic_bias};
\cite{BR_Net_dataset_vs_task};
\cite{minority_group_vs_sensitive_attribute};
\cite{spurious_correlation_Underrepresentation}.

\subsubsection{Fairness criteria}
\cite{FAMS};
\cite{FAIRREPROGRAM};
\cite{Knowledge_Distillation};
\cite{MIP};
\cite{fairness_distribution_shift};
\cite{IAF};
\cite{topdown};
\cite{Information_projection};
\cite{Fair_IJ};
\cite{fairer};
\cite{fairness_calibration};
\cite{EL};
\cite{FBDE};
\cite{differential_privacy};
\cite{correlation_shift};
\cite{DARTS};
\cite{fairDRO};
\cite{fifa};
\cite{FNF};
\cite{strategic_manipulation_fairness};
\cite{Data_reweighting_with_influence};
\cite{instance_level_fairness_constraints};
\cite{adaptive_weights};
\cite{FairSmooth};
\cite{RNF};
\cite{sample_selection};
\cite{Logistic_Fairness_Relaxations};
\cite{retiring_adult};
\cite{fairwashing};
\cite{FairAdj};
\cite{mixup};
\cite{fairbatch};
\cite{DLR};
\cite{Ideal};
\cite{GroupFair};
\cite{meta_algorithm};
\cite{KDE};
\cite{random_perturbation};
\cite{fact};
\cite{Private_Demographic_Data};
\cite{saha2020measuring};
\cite{lohaus2020too};
\cite{Zhao2020Conditional};
\cite{Baharlouei2020Renyi};
\cite{williamson2019fairness};
\cite{chzhen2019leveraging};
\cite{lamy2019noise};
\cite{LAFTR};
\cite{pmlr-v80-liu18c};
\cite{pmlr-v80-kilbertus18a};
\cite{pmlr-v80-kearns18a};
\cite{pmlr-v80-agarwal18a};
\cite{NIPS2017_e6384711};
\cite{orthogonal};
\cite{Fairalm_DP_difference_false_positive_rate};
\cite{fairnessgan_DP_difference_error_rate};
\cite{EO_refine};
\cite{gradient_projection};
\cite{LFR};
\cite{fairgan+};
\cite{richardson2023add};
\cite{bell2023possibility};
\cite{defrance2023maximal};
\cite{calvi2023enhancing};
\cite{ganesh2023impact};
\cite{petersen2023assessing};
\cite{alvarez2023domain};
\cite{almuzaini2022abcinml};
\cite{black2022algorithmic};
\cite{baumann2022enforcing};
\cite{mishler2021fade};
\cite{grabowicz2022marrying};
\cite{zhang2022affirmative};
\cite{kong2022intersectionally};
\cite{sikdar2022getfair};
\cite{pfohl2022net};
\cite{agarwal2022power};
\cite{sharaf2022promoting};
\cite{singh2021fairness};
\cite{raz2021group};
\cite{rodolfa2020case};
\cite{slack2020fairness};
\cite{liu2020disparate};
\cite{harrison2020empirical};
\cite{kallus2022assessing};
\cite{celis2019classification};
\cite{friedler2019comparative};
\cite{menon2018cost};
\cite{beckerstandardized};
\cite{sharma2023far};
\cite{tifreafrappe};
\cite{xuadapting};
\cite{schrouff2024mind};
\cite{zhangtowards};
\cite{pang2025fairness};
\cite{taufiqachievable};
\cite{luo2024your};
\cite{wang2024removing};
\cite{chen2024posthoc};
\cite{grari2024on};
\cite{chowdhury2024enhancing};
\cite{chen2024posthoc};
\cite{yin2024fair};
\cite{liu2024empirical};
\cite{tian2024fairseg};
\cite{weerts2024neutrality};
\cite{zezulka2024fair};
\cite{poe2024conflict};
\cite{ni2024fairness};
\cite{chan2024group};
\cite{somerstep2024algorithmic};
\cite{laszkiewicz2024benchmarking};
\cite{gillis2024operationalizing};
\cite{jaime2024ethnic};
\cite{blandin2024learning};
\cite{rateike2024designing};
\cite{wyllie2024fairness};
\cite{mhasawade2024causal}.

\subsection{Survey about bias issues}
\cite{fairness_survey_DP_EO};
\cite{bias_recommender_system};
\cite{prediciton_quality_disparity};
\cite{discussion_on_DP_EO};
\cite{causal_based_fairness};
\cite{MLbias_survey};
\cite{bias_in_visual_datasets};
\cite{datasets_ML};
\cite{protected_attributes};
\cite{fairness_criminal_justice};
\cite{critical_bias_survey_NLP};
\cite{benbouzid2023fairness};
\cite{devinney2022theories};
\cite{schwobel2022long};
\cite{finocchiaro2021bridging};
\cite{hutchinson201950};
\cite{binns2018fairness};
\cite{baumann2023bias};
\cite{AIF360};
\cite{hort2022bia}
\cite{caton2020fairness};
\cite{delaney2024oxonfair};
\cite{jin2024fairmedfm};
\cite{buyl2024fairret};
\cite{han2024ffb};
\cite{deck2024critical}.

\subsection{Bias assessment metrics}
\cite{GDP};
\cite{BA};
\cite{Directional_BA};
\cite{multi_BA};
\cite{model_leakage};
\cite{divergence_between_score_distributions};
\cite{RLB};
\cite{disparate_mistreatment_on_FPR};
\cite{SensitiveNets};
\cite{FlowSAN};
\cite{DP_FFVAE};
\cite{distance_correlation};
\cite{CAT};
\cite{fairness_constraints};
\cite{Resound} 
\cite{Repair};
\cite{CAI_rz};
\cite{hiranandani2020fair};
\cite{leino2018feature}.

\subsection{Fairness constraints}
\cite{EO_define};
\cite{demographic_parity};
\cite{counterfactual_fairness};
\cite{fairness_through_awareness};
\cite{fairness_under_unawareness};
\cite{discrimination_score};
\cite{impossibility_for_fair_repesentations};
\cite{process_fairness_FPR_difference};
\cite{AIF360};
\cite{reject_option_classification};
\cite{certified_fairness};
\cite{bechavod2024monotone};
\cite{munagala2024individual};
\cite{xu2024intersectional}.

\end{document}